\documentclass[10pt,twocolumn,letterpaper]{article}

\usepackage[pagenumbers]{cvpr} 

\usepackage{times}
\usepackage{epsfig}
\usepackage{subcaption}
\usepackage{graphicx}
\graphicspath{{./Figures/}}
\usepackage{amsmath}
\usepackage{amssymb}
\usepackage{amsthm}
\usepackage{algcompatible}
\usepackage{algorithm}
\usepackage[noend]{algpseudocode}

\usepackage[pagebackref,breaklinks,colorlinks]{hyperref}

\usepackage[capitalize]{cleveref}
\crefname{section}{Sec.}{Secs.}
\Crefname{section}{Section}{Sections}
\Crefname{table}{Table}{Tables}
\crefname{table}{Tab.}{Tabs.}

\begin{document}

\title{Fast geometric trim fitting using\\partial incremental sorting and accumulation}

\author{Min Li\\
ShanghaiTech University\\
{\tt\small limin1@shanghaitech.edu.cn}
\and
Laurent Kneip\\
ShanghaiTech University\\
{\tt\small lkneip@shanghaitech.edu.cn}
}
\maketitle

\begin{abstract}
   We present an algorithmic contribution to improve the efficiency of robust trim-fitting in outlier affected geometric regression problems. The method heavily relies on the quick sort algorithm, and we present two important insights. First, partial sorting is sufficient for the incremental calculation of the $x$-th percentile value. Second, the normal equations in linear fitting problems may be updated incrementally by logging swap operations across the $x$-th percentile boundary during sorting. Besides linear fitting problems, we demonstrate how the technique can be additionally applied to closed-form, non-linear energy minimization problems, thus enabling efficient trim fitting under geometrically optimal objectives. We apply our method to two distinct camera resectioning algorithms, and demonstrate highly efficient and reliable, geometric trim fitting.
\end{abstract}

\section{Introduction}

Over the past decade, deep convolutional neural networks (CNNs) have lead to significant advances in automated visual perception capabilities. However, it remains true that the performance of CNNs in classification problems still dominates regression performance despite the availability of large amounts of training data. Accurate model-based geometric fitting therefore remains important, especially if the dimensionality of the problem is low and thus tractable via commonly available solvers. The present paper addresses a very classical problem in geometric fitting, which is outlier contamination in the data.
\begin{figure}[t!]
\setlength{\abovecaptionskip}{+0.1cm}
\setlength{\belowcaptionskip}{-0.4cm}
\begin{subfigure}{0.5\textwidth}
\centering
\includegraphics[width=1\textwidth]{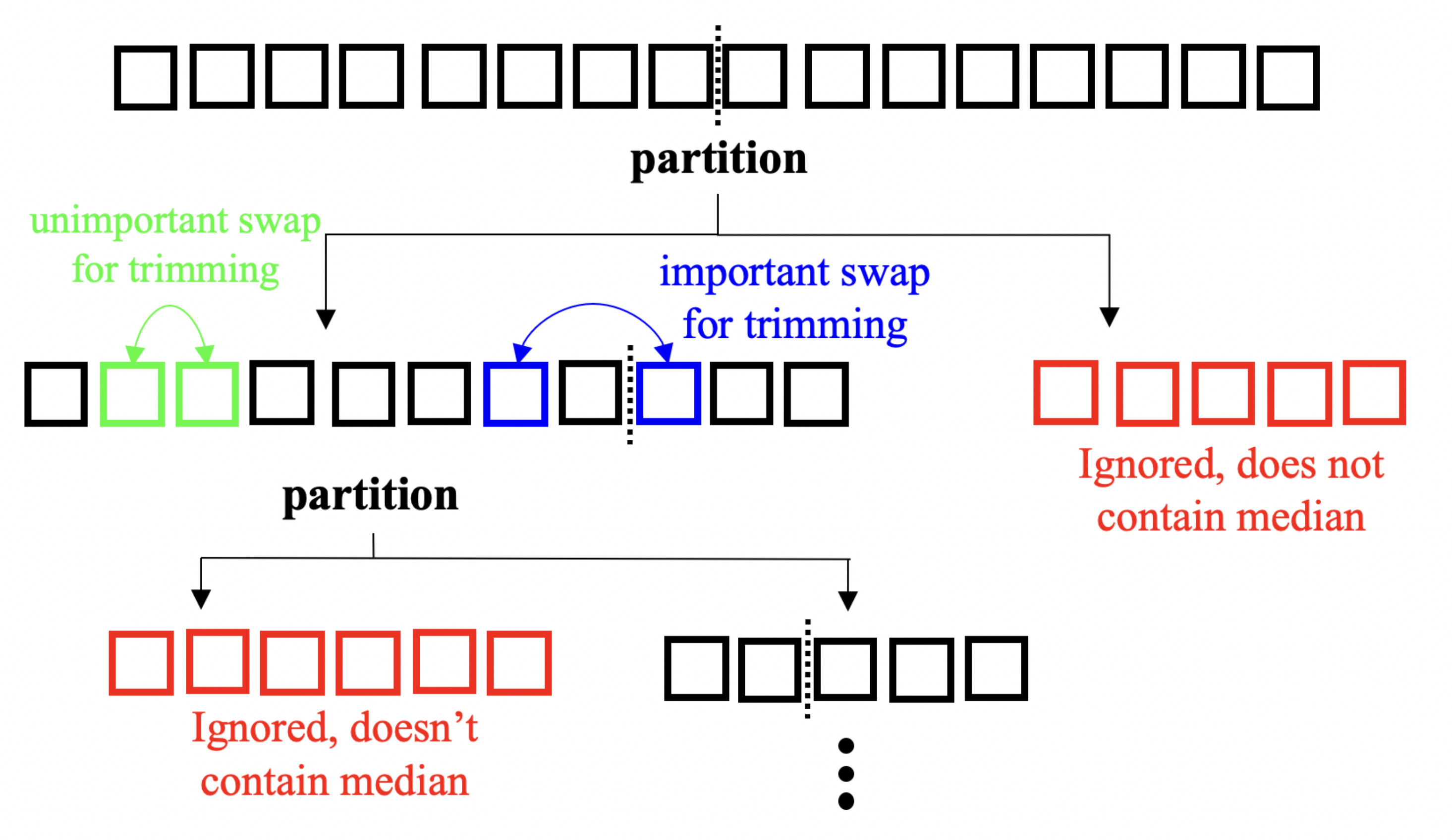}
\end{subfigure}

\begin{subfigure}{0.24\textwidth}
\flushleft
\includegraphics[width=1\textwidth]{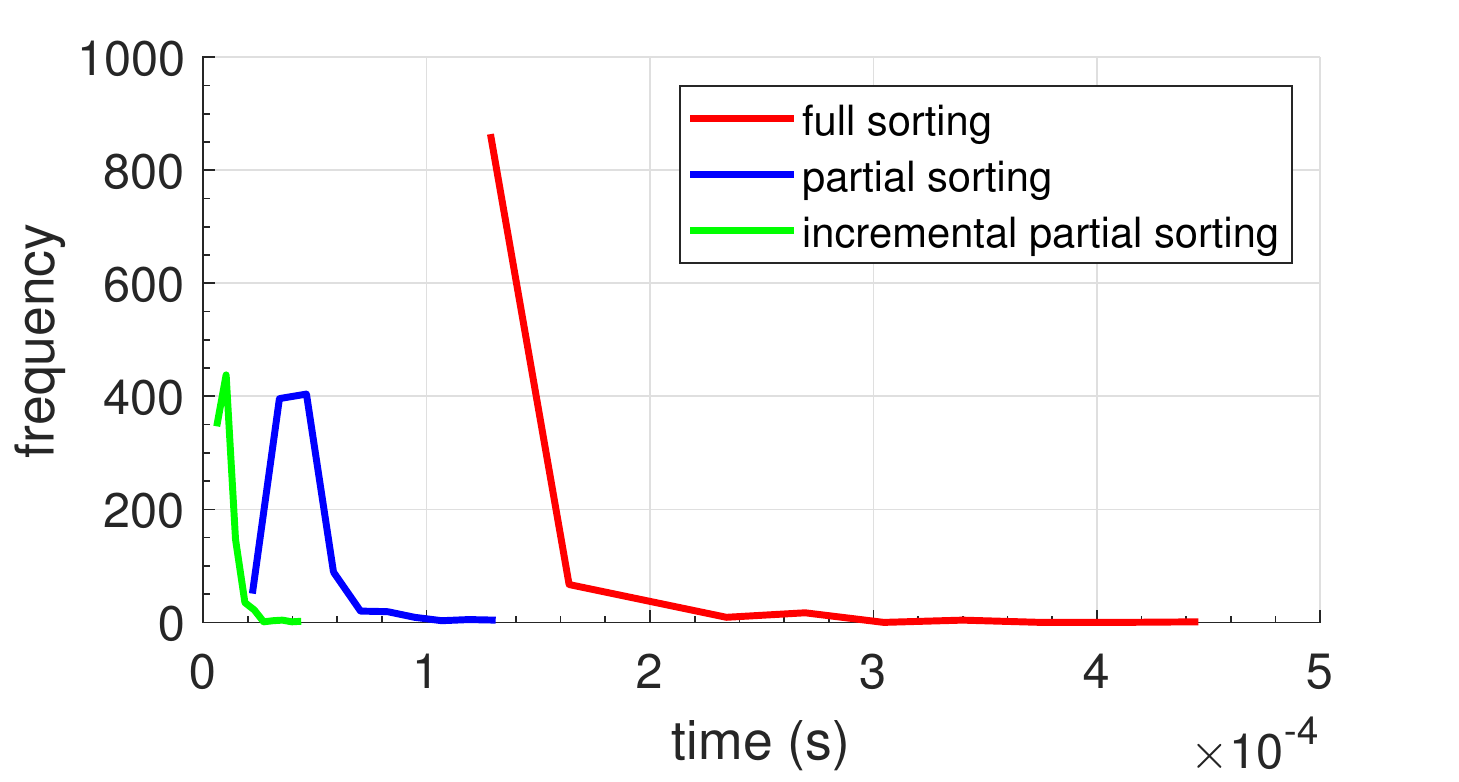}
\end{subfigure}
\hspace{-0.2cm}
\begin{subfigure}{0.24\textwidth}
\flushleft
\includegraphics[width=1\textwidth]{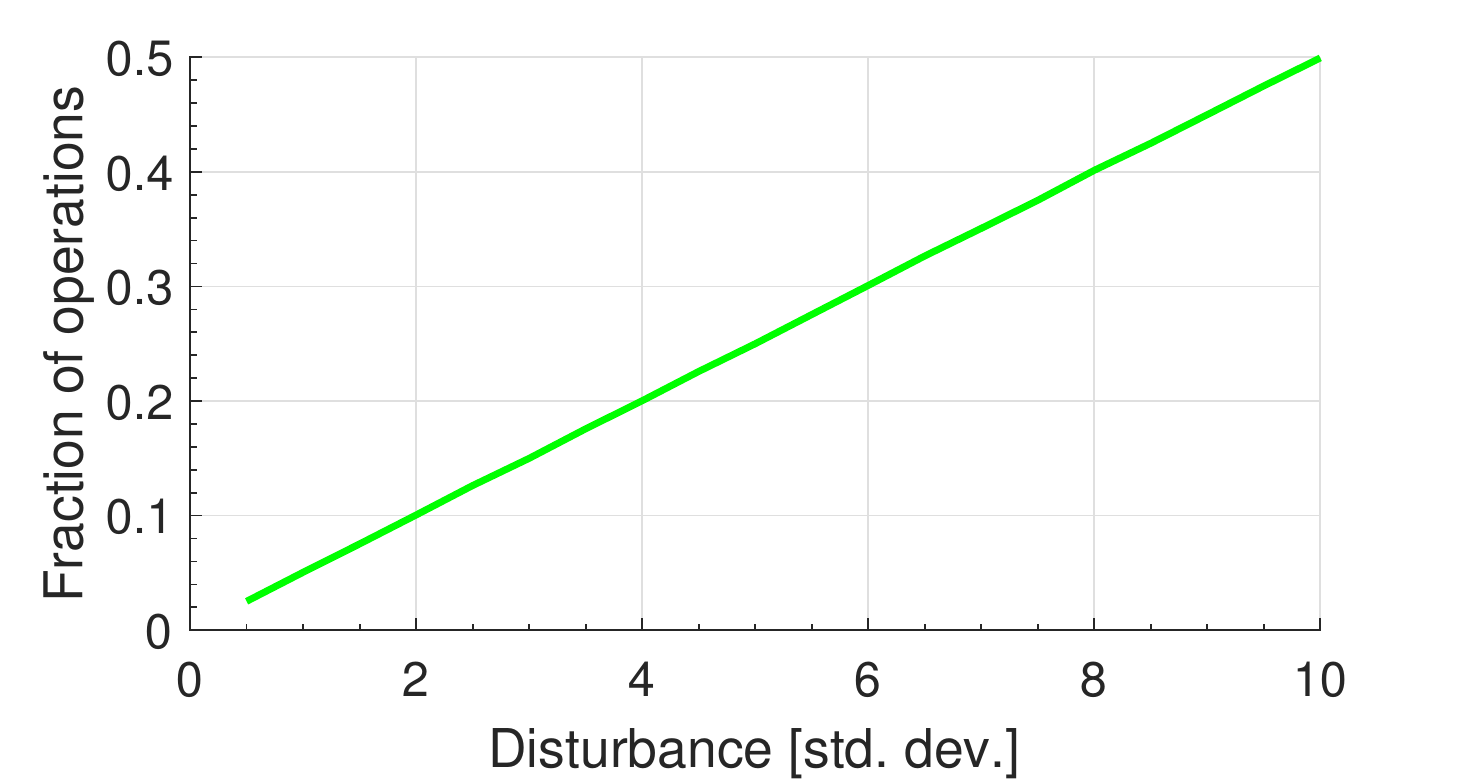}
\end{subfigure}

\caption{Illustration of our sorting for trimming strategy. Only the part that contains the trimming boundary needs to be further processed, and only swaps across the trimming boundary are important for further processing. Bottom left: Histogram over times for sorting an array of doubles. Incremental sorting is applied to a pre-sorted array for which the values are disturbed. Bottom right: Incremental summation of elements smaller than the median after value perturbation and incremental sorting. The plot shows the fraction of operations required with respect to naive summation.}
\label{fig:Figure1}
\end{figure}

Solutions to robust geometric fitting may be grouped into two sub-categories. The first one is given by bottom-up approaches such as the popular RANSAC~\cite{fischler81} algorithm. The second class of algorithms are top-down methods which try to directly fit the model to all the data by using a robust cost function~\cite{aftab15} or global optimization~\cite{enqvist08}. However, robust kernel based methods depend on a sufficiently good initial guess, and global optimization methods are often computationally demanding and thus only work for low dimensions.

An interesting top-down alternative is given by robust nullspace estimation methods such as \textit{dual principal component pursuit} proposed by Tsakiris et al.~\cite{tsakiris18}. While the method is showing advantages over RANSAC in situations in which the subspace dimension is large, it remains computationally demanding and often cannot be considered as a viable alternative in real-time applications. The present work picks up the work by Ferraz et al.~\cite{ferraz14} which introduces a very fast, trimming-based technique for robust nullspace fitting. Though the method is similar than many others in that it does not guarantee optimality of the identified outlier subset, we demonstrate that it works surprisingly well in the real-world application of camera resectioning.

Our work makes two important contributions:
\begin{itemize}
  \item We introduce two important algorithmic modifications to the commonly applied sorting mechanisms in trimming approaches. We demonstrate that partial, incremental sorting is enough. We furthermore demonstrate how the swapping nature of common sorting algorithms can be immediately reused to improve the efficiency of the iterative solution of the null-space.
  \item Robust linear null-space fitting typically employs an algebraic error. We extend the idea of robust trim fitting from linear null-space fitting to geometrically optimal, non-linear closed-form solvers.
\end{itemize}

In the practical part of our work, we show an application of the idea to robust camera resectioning, demonstrating outstanding computational efficiency and success rate even in challenging, high-outlier scenarios.

\section{Related work}
\label{sec:related}

Geometric fitting problems often appear in computer vision and aim at solving the absolute camera pose resectioning problem~\cite{lepetit09,hesch11,li12,zheng13,kneip14}, the relative camera pose problem~\cite{hartley97}, homography estimation for pure rotation and planar structure~\cite{sutherland74,hartley04}, or 3D point set registration~\cite{arun87,horn87,umeyama91}. For absolute and relative camera pose problems, there also exist minimal~\cite{kneip11,nister04b} and directional-correspondence-based~\cite{kukelova10,fraundorfer10} solvers. The body of literature on geometric problems is large and the algorithms listed here are only some of the more established solvers for a few of the more fundamental problems. The present work addresses the solution of geometric fitting problems in the presence of outlier samples.

A popular way to deal with outliers consists of moving from a least-squares estimator to the more general class of M-estimators~\cite{hayashi00}. As originally demonstrated by Weiszfeld~\cite{weiszfeld37}, increased robustness against outliers can be obtained by exchanging the common least-squares $L_2$-norm objective against the $L_q$-norm objective (with $q$ smaller than 2)\footnote{The $L_q$ Weiszfeld algorithm seeks the $L_q$ mean which minimizes the the sum of the $q$-th power of the residual distances for given samples}. Aftab and Hartley~\cite{aftab15} furthermore prove that Iteratively Reweighted Least Squares (IRLS) for properly chosen weights converges to the $L_q$ Weiszfeld algorithm. Furthermore, they prove that IRLS can be extended to an entire family of robust M-estimators that employ robust norms and kernels (e.g. Huber norm, Pseudo-Huber norm)~\cite{hartley04}. While the application of robust kernels and IRLS is an established technique, the method employs local gradients and depends on the availability of a sufficiently good initial guess.

In order to achieve optimal identification of outliers, the community has proposed a number of globally optimal solutions to inlier cardinality maximization~\cite{enqvist08,ask13,enqvist15,svarm06,sim06,yang14,liu21}. These methods often employ the $L_{\infty}$-norm and utilize the branch-and-bound algorithm. While certainly interesting from a theoretical stand-point, the branch-and-bound algorithm suffers from the curse of dimensionality and quickly becomes computationally demanding as the dimensionality of the problem increases. Solutions based on branch-and-bound are often computationally intractable in real-time applications.

A more efficient and established technique for dealing with outliers in geometric fitting problems is given by the RANSAC~\cite{fischler81} algorithm. It typically employs a solver that finds an initial hypothesis for the model parameters from a minimal, randomly sampled set of input samples. In an alternating second step, the scheme then aims at determining the amount of consensus between this hypothesis and all other sample points. The procedure is repeated until convergence. Several extensions to the algorithm have been proposed over the years (MLESAC~\cite{torr00}), Preemptive RANSAC~\cite{nister05}, PROSAC~\cite{chum05}, GroupSAC~\cite{ni09}). While the RANSAC algorithm still counts as the standard solution to robust geometric fitting, the success of the algorithm is compromised for larger cardinalities of the minimal sample set required to establish a hypothesis.

Recent times have seen the surge of competing, powerful top-down algorithms that may even come with convergence guarantees. Tsakiris and Vidal~\cite{tsakiris18} present the DPCP algorithm for robust nullspace fitting, which works with a dual representation of the nullspace and aims at finding its orthogonal complement. The approach works well particularly in scenarios where the sub-space dimension is large compared to the ambient dimension, a situation in which RANSAC often fails. The success of the method is recently demonstrated by Ding et al.~\cite{ding20}, who successfully apply the method to homography estimation problems. In 2014, Ferraz et al.~\cite{ferraz14} propose a highly efficient alternative to the IRLS algorithm. Rather than performing iterative reweighting of the samples, the algorithm performs trimming and iteratively solves for the nullspace using the $n$-th percentile of samples sorted by their residual distances. The algorithm works for linear problems, which is both an advantage and a disadvantage. The advantage is that no gradients are required and the method operates in closed-form. The disadvantage is that it often implies the use of simplified algebraic cost functions. Our work makes two important contributions with respect to this technique. First, we introduce important insights that lead to a significant algorithm speed-up of this and in fact all trim fitting approaches. Second, we demonstrate that the idea remains amenable to non-linear, geometrically optimal closed-form solvers.

\section{Theory}

We formulate our theory from an abstract perspective. Let
$\mathbf{y}=\mathbf{f}_{\boldsymbol{\theta}}(\mathbf{x})$
be a vectorial function $\mathbf{f}:\mathbb{R}^{n}\rightarrow\mathbb{R}^{m}$ that depends on the parameter vector $\boldsymbol{\theta}$. Our approach aims at fitting problems which---in their most basic form---have an input given by a set of $N$ noisy input correspondences
\begin{equation}
    \mathcal{S} = \left\{ \left\{ \mathbf{\tilde{x}}_{1}, \mathbf{\tilde{y}}_{1} \right\}, \ldots, \left\{ \mathbf{\tilde{x}}_{N}, \mathbf{\tilde{y}}_{N} \right\} \right\}\text{, }\mathbf{\tilde{x}}_i\in\mathbb{R}^{n} \text{ and } \mathbf{\tilde{y}}_i\in\mathbb{R}^{m}.
    \label{eq:correspondences}
\end{equation}
The problem consists of identifying the optimal parameters $\boldsymbol{\theta}^*$ such that $E = \sum_{i=1}^{N} \| \mathbf{\tilde{y}}_{i} - \mathbf{f}_{\boldsymbol{\theta}^*}(\mathbf{\tilde{x}}_{i}) \|^2$ is minimized. The function $\mathbf{f}_{\boldsymbol{\theta}}$ is often a non-linear function and numerous algebraic linearizations of such non-linear functions have been introduced. A more general form of our objective is therefore given by minimizing
\begin{equation}
    \boldsymbol{\theta}^* = \underset{\boldsymbol{\theta}}{\operatorname{argmin}}\sum_{i=1}^{N} \| \mathbf{r}(\mathbf{\tilde{x}}_{i},\mathbf{\tilde{y}}_{i},\boldsymbol{\theta}) \|^2,
\end{equation}
where $\mathbf{r}(\cdot)$ represents a residual function that vanishes for any parameter $\boldsymbol{\theta}$ that brings $\mathbf{\tilde{x}}_{i}$ into ideal agreement with $\mathbf{\tilde{y}}_{i}$. One of the two following statements often holds:
\begin{itemize}
    \item $\mathbf{r}(\cdot)$ is linear in $\boldsymbol{\theta}$ and the entire objective therefore can be solved using linear least squares. However, the residual error does not correspond to a clearly defined, geometric distance, and it is non-trivial to make a statement about the optimality of the identified solution.
    \item $\mathbf{r}(\cdot)$ is scalar-valued and corresponds to a clearly defined geometric distance, but is non-linear in nature. The objective may therefore only be solved using non-linear least-squares solvers or---in the situation of a polynomial form---Gr\"obner basis solvers derived for the first-order optimality conditions.
\end{itemize}

The above only outlines the most basic form of the problem in which the correspondence set $\mathcal{S}$ is not affected by \textit{outliers} (e.g. correspondences which are not following assumptions made by a Gaussian noise model). Let $\mathcal{S}_{in}\subset\mathcal{S}$ be the subset of maximum cardinality for which parameters $\boldsymbol{\theta}^*$ exist such that
\begin{equation}
    \|\mathbf{r}(\mathbf{\tilde{x}}_{i},\mathbf{\tilde{y}}_{i},\boldsymbol{\theta}^*)\| < \tau \text{ }\forall \left\{\mathbf{\tilde{x}}_{i},\mathbf{\tilde{y}}_{i}\right\} \in \mathcal{S}_{in}.
\end{equation}
Outlier robust fitting is therefore often formulated as a cardinality maximization problem over $\boldsymbol{\theta}$
\begin{equation}
    \boldsymbol{\theta}^* = \underset{\boldsymbol{\theta}}{\operatorname{argmax}}\sum_{i=1}^{N} \delta( \|\mathbf{r}(\mathbf{\tilde{x}}_{i},\mathbf{\tilde{y}}_{i},\boldsymbol{\theta})\| < \tau ),
\end{equation}
where $\delta(\cdot)$ is the indicator function and returns one if the internal condition is true, and zero otherwise. $\mathcal{S}_{out} = \mathcal{S} \setminus \mathcal{S}_{in}$ is defined as the set of outliers, and $\tau$ is a pre-defined inlier threshold. As mentioned in Section \ref{sec:related}, many approaches to this problem have already been presented. In the following, we will introduce a very fast, robust trim fitting approach.

\subsection{Trimming using partial incremental sorting}

Similar to the REPPnP method~\cite{ferraz14}, the core of our robust geometric fitting algorithm is given by a trimming strategy in which samples are ranked by how well they fit a hypothesis $\boldsymbol{\theta}_k$ in terms of the residual error $\|\mathbf{r}(\mathbf{\tilde{x}}_{i},\mathbf{\tilde{y}}_{i},\boldsymbol{\theta}_k)\|$. The $x$-th percentile of the data is then alternatingly used to calculate new model parameters $\boldsymbol{\theta}_{k+1}$.

Formally, the algorithm consists of the alternating execution of two steps. In step one, we use the current hypothesis $\boldsymbol{\theta}_k$ to obtain the sorted set of correspondences
\begin{eqnarray}
    \mathbf{s}_k & = & \left[ \left\{\mathbf{\tilde{x}}_{j_1},\mathbf{\tilde{y}}_{j_1}\right\},\ldots,\left\{\mathbf{\tilde{x}}_{j_N},\mathbf{\tilde{y}}_{j_N}\right\}\right]\text{, where} \nonumber \\
    a < b & \Leftrightarrow & \|\mathbf{r}(\mathbf{\tilde{x}}_{j_a},\mathbf{\tilde{y}}_{j_a},\boldsymbol{\theta}_k)\|\leq\|\mathbf{r}(\mathbf{\tilde{x}}_{j_b},\mathbf{\tilde{y}}_{j_b},\boldsymbol{\theta}_k)\|.
\end{eqnarray}
The generation of $\mathbf{s}_k$ obviously requires the execution of a sorting algorithm. Step two then consists of finding new model parameters with the $x$-th percentile of lowest residual correspondences, i.e.
\begin{equation}
    \boldsymbol{\theta}_{k+1} = \underset{\boldsymbol{\theta}}{\operatorname{argmin}}\sum_{i=1}^{\lfloor\frac{x}{100}N\rfloor} \| \mathbf{r}(\mathbf{\tilde{x}}_{j_i},\mathbf{\tilde{y}}_{j_i},\boldsymbol{\theta}) \|^2.
    \label{eq:trimFitting}
\end{equation}

Our first main contribution relies on the insight that---in case of using a fixed $x$-th percentile---only partial sorting of $\mathbf{s}$ is required. We use the \textit{quick sort} divide-and-conquer algorithm, for which the main steps are as follows:
\begin{itemize}
    \item Pick one element in the set as pivot element.
    \item Partition the remaining elements into two sub-sets such that any element in subset one is smaller than the pivot element, and any element in sub-set two is larger than the pivot element. The partition algorithm works by using two indices $\alpha$ and $\beta$ that scan the array from the smallest to the largest element and vice-versa. Incrementing $\alpha$ is paused as soon as an element bigger than the pivot element is encountered. Decrementing $\beta$ is paused as soon as an element smaller than the pivot element is encountered. The two elements are swapped, and the scanning continues. As soon as the indices cross, the partitioning is finished. The pivot element is placed at the boundary by another swapping operation. This concludes the partitioning with the desired property.
    \item Recursively apply to both sub-sets.
\end{itemize}

This is text-book knowledge, so further details are omitted here. The important insight is that recursive application can be limited to one of the subsets, only. Let us denote by $p$ the final position of the pivot element in $\mathbf{s}_{k}$ after the first partitioning step is completed. The position of the pivot element segments $\mathbf{s}_{k}$ into a left part $\mathbf{s}_{kl}$ and a right part $\mathbf{s}_{kr}$ such that any element in $\mathbf{s}_{kl}$ is less or equal to the pivot element, and any element in $\mathbf{s}_{kr}$ is bigger than the pivot element (note furthermore that every element in $\mathbf{s}_{kl}$ is smaller than any of the elements from $\mathbf{s}_{kr}$). Three scenarios may occur:
\begin{itemize}
    \item $p>\lfloor\frac{x}{100}N\rfloor$: In this case, only the set $\mathbf{s}_{kl} = \left[\left\{ \tilde{\mathbf{x}}_{i_1}, \tilde{\mathbf{y}}_{i_1} \right\},\ldots,\left\{ \tilde{\mathbf{x}}_{i_p}, \tilde{\mathbf{y}}_{i_p} \right\}\right]$ needs further processing.
    \item $p<\lfloor\frac{x}{100}N\rfloor$: In this case, only the set $\mathbf{s}_{kr}=\left[\left\{ \tilde{\mathbf{x}}_{i_{p+1}}, \tilde{\mathbf{y}}_{i_{p+1}} \right\},\ldots,\left\{ \tilde{\mathbf{x}}_{i_N}, \tilde{\mathbf{y}}_{i_N} \right\}\right]$ needs further processing.
    \item $p=\lfloor\frac{x}{100}N\rfloor$: In this case, the algorithm may be readily terminated. The $x$-th percentile score is given by $\|\mathbf{r}(\tilde{\mathbf{x}}_{i_p},\tilde{\mathbf{y}}_{i_p},\boldsymbol{\theta_k})\|$.
\end{itemize}

It is furthermore clear that---as the overall fitting algorithm approaches convergence---an increasing number of correspondences that have been ranked within the $x$-th percentile eventually remain within that percentile even after an update has been generated. If starting from the previous order, only a limited number of swapping operations will need to be executed. A small experiment in which we simply calculate the median (i.e. 50-th percentile) of an array of numbers is presented in Figure \ref{fig:Figure1}. Numbers are uniformaly sampled from the interval $[-10,10]$, and results are averaged over 1000 experiments. As expected, partial sorting significantly increases the computational efficiency of retrieving the median. If we sort the vector, add a perturbation to the elements by uniformly sampling in the interval $[-1,1]$, and then repeat the sorting, another substantial gain in computational efficiency can be observed. The complexity of the median retrieval behaves approximately linear in the number of points.

\subsection{Incremental accumulation}

Next, let us suppose that the residual may be written in polynomial form. We have
\begin{equation}
    \mathbf{r}(\tilde{\mathbf{x}}_i,\tilde{\mathbf{y}}_i,\boldsymbol{\theta}) = \mathbf{D}(\tilde{\mathbf{x}}_i,\tilde{\mathbf{y}}_i) \cdot \mathbf{m}(\boldsymbol{\theta}),
\end{equation}
where $\mathbf{D}(\tilde{\mathbf{x}}_i,\tilde{\mathbf{y}}_i)$ is a matrix that depends only on the data, and $\mathbf{m}(\boldsymbol{\theta})$ is a column vector that consists of different monomial forms of the unknowns. Note that linear forms are included in polynomial forms and simply given if $\mathbf{m}(\boldsymbol{\theta})=\boldsymbol{\theta}$. Given this form, the objective that needs to be updated in each iteration can be written as
\small
\begin{eqnarray}
    \boldsymbol{\theta}_{k+1} &=& \underset{\boldsymbol{\theta}}{\operatorname{argmin}} \text{ }\mathbf{m}(\boldsymbol{\theta})^T \left[ \sum_{i=1}^{\lfloor\frac{x}{100}N\rfloor} \mathbf{D}(\tilde{\mathbf{x}}_i,\tilde{\mathbf{y}}_i)^T \mathbf{D}(\tilde{\mathbf{x}}_i,\tilde{\mathbf{y}}_i) \right] \mathbf{m}(\boldsymbol{\theta})\nonumber \\
    &=& \underset{\boldsymbol{\theta}}{\operatorname{argmin}} \text{ }\mathbf{m}(\boldsymbol{\theta})^T \cdot \mathcal{A}_k \cdot \mathbf{m}(\boldsymbol{\theta}).
\end{eqnarray}
\normalsize
$\mathcal{A}_k$ is what we denote here as an \textit{accumulator}. It is the accumulator in the $k$-th iteration which is composed by using the $x$-th percentile of data given the sorting in that iteration. Note that many if not the majority of non-minimal fitting algorithms include a similar formation of an accumulator as one of their sub steps. In both linear as well as iteratively linearized non-linear regression problems, the formation of the present accumulator is what we do when forming the normal equations of the system, and a solution or update is found by singular value decomposition of $\mathcal{A}_k$. In closed-form non-linear solvers, the elements of $\mathcal{A}_k$ are used to fill the elimination template of a Gr\"obner basis solver.

The second main contribution of our fast trimming strategy then relies on the insight that the accumulator $\mathcal{A}_k$ does not have to be recalculated in each iteration. More specifically, since the quick sort algorithm performs sorting by a sequence of swapping operations, the accumulator can be incrementally updated whenever we swap a pair of correspondences for which one is on the left side of the $x$-th percentile boundary, and the other one on the right. Given that the number of such swap operations in partial incremental sorting is substantially lower than the number of actual correspondences, we again obtain a significant gain in computational efficiency. We denote our algorithm \textit{quicksort4trim}, and it is defined to return two logs denoted \textit{plusLog} and \textit{minusLog}. The latter refer to the indices of the elements involved in cross-percentile-boundary swaps during sorting, and thus have to be added or removed from the accumulator (note that redundant swaps are ignored). The effectiveness of this approach is again verified in a small experiment in which we incrementally calculate the sum of all elements smaller than the median. As indicated in Figure \ref{fig:Figure1}, the number of required operations to update the sum for moderate perturbations can be as low as 10\% of the number of operations required for a naive summation. Accumulators in geometric fitting often involve matrix operations, which is why the impact on overall computational efficiency can be substantial.

\section{Application to camera resectioning}

We apply our robust trim fitting strategy to a classical problem from geometric computer vision: \textit{camera pose resectioning}. After a definition of the problem, we will first see an efficient, incremental variant of the original \textit{REPPnP} algorithm proposed by Ferraz et al.\cite{ferraz14}. We will furthermore see an application of the incremental trimming strategy to a geometrically optimal, closed-form non-linear solver, which is the \textit{UPnP} algorithm by Kneip et al.~\cite{kneip14}.

\subsection{Problem statement}

The goal of the Perspective-$n$-Point (P$n$P) problem is to find extrinsic camera pose parameters $\mathbf{R}$ and $\mathbf{t}$ that transform points $\mathbf{p}_i$ from the world frame to the camera frame such that they come into alignment with direction vectors $\mathbf{f}_{i}$ measured in the camera frame, i.e.
\begin{equation}
    \lambda_i \mathbf{f}_{i} = \mathbf{R}\mathbf{p}_i+\mathbf{t}.
\end{equation}
$\lambda_i$ denotes the unknown depth of the point seen from the camera frame. The P$n$P problem is solved for an arbitrarily large number of points, and state-of-the-art solutions typically have linear complexity in this number.

\subsection{Incremental REPPnP}

\textit{REPPnP} by Ferraz et al.~\cite{ferraz14} is strongly inspired by the \textit{EPnP} algorithm~\cite{lepetit09} and relies on the prior extraction of control points in the world frame. Using the latter, every world point can be expressed as a linear combination $\mathbf{p}_i^w = \sum_{j=1}^{4} \alpha_{ij} \mathbf{c}_j^w$. Knowing that the linear combination weights do not depend on the reference frame, it is easy to see that
\begin{equation}
    \mathbf{p}_i^c = \lambda_i \mathbf{f}_{i} = \sum_{j=1}^{4} \alpha_{ij} \mathbf{c}_j^c.
\end{equation}
Assuming that $\mathbf{f}_i = \left[u_i^c\text{ }v_i^c\text{ }1\right]^T$, the third row can be used to eliminate the unknown depth, and it immediately follows that
\begin{eqnarray}
    \left[\begin{matrix}\alpha_{i1} & \alpha_{i2} & \alpha_{i3} & \alpha_{i4}\end{matrix}\right]\otimes \left[\begin{matrix}1 & 0 & -u_i^c \\ 0 & 1 & -v_i^c\end{matrix}\right] \boldsymbol{\theta} &=& \mathbf{0}\nonumber\\
    \Leftrightarrow \mathbf{D}_i \boldsymbol{\theta} &=& \mathbf{0}
\end{eqnarray}
where $\boldsymbol{\theta}^T = \left[\begin{matrix} \mathbf{c}_1^{cT} &  \mathbf{c}_2^{cT} & \mathbf{c}_3^{cT} & \mathbf{c}_4^{cT} \end{matrix} \right]$ is the solution space given by the control points expressed in the camera frame, and $\otimes$ denotes the Kronecker product. The camera pose is subsequently found by control point alignment.

\textit{REPPnP} solves this problem robustly via trim fitting. For $N$ points, it iteratively updates $\boldsymbol{\theta}$ by nullspace extraction, i.e.
\begin{equation}
    \boldsymbol{\theta} \leftarrow \mathcal{NS}\left(\left[ \sum_{i=1}^{N} w_i \mathbf{D}_{i}^T \mathbf{D}_i \right]\right).
\end{equation}
Originally, $w_i=1 \forall i$. Let $\boldsymbol{\theta}_k$ be the solution found in iteration $k$. The $w_i$ are then updated such that $w_i=1$ if $\|\mathbf{D}_i\boldsymbol{\theta}_k\|<\tau$, and $w_i=0$ otherwise. $\tau$ is defined as the median of the sequence
\begin{equation}
    \mathbf{s} = \left[ \|\mathbf{D}_1\boldsymbol{\theta}_k\|,\ldots,\|\mathbf{D}_N\boldsymbol{\theta}_k\|\right].
\end{equation}

The original \textit{REPPnP} algorithm applies full sorting and accumulation in each iteration. The incremental version of \textit{REPPnP}---denoted \textit{REPPnPIncr}---is obtained by applying our \textit{quicksort4trim} partial sorting algorithm and performing incremental accumulation. We use the 50-th percentile throughout this paper, and the resulting algorithm is summarized in Algorithm \ref{alg:incrreppnp}.

\begin{algorithm}[t]
\caption{Incremental REPPnP}
\label{alg:incrreppnp}
\begin{algorithmic}[1]

\Procedure{REPPnPincr}{$[\{\mathbf{f}_1,\mathbf{p}_1\},\ldots,\{\mathbf{f}_N,\mathbf{p}_N\}]$}
    \State calculate control points and combination weights
    \State extract $[\mathbf{D}_1,\ldots,\mathbf{D}_N]$
    \State $\mathcal{A} = \sum_{i=1}^{N} \mathbf{D}_i^T\mathbf{D}_i$
    \State $\boldsymbol{\theta}_0 \leftarrow \mathcal{NS}(\mathcal{A})$ \Comment{$\mathcal{NS}(\cdot)$ extracts nullspace}
    \State init $\mathbf{s}=[\{\|\mathbf{D}_1\boldsymbol{\theta}_0\|,1\},\ldots,\{\|\mathbf{D}_N\boldsymbol{\theta}_0\|,N\}]$
    \State $\mathcal{A}=\sum_{i=1}^{\lfloor N/2\rfloor}\mathbf{D}_i^T\mathbf{D}_i$

    \While{not converged}
        \State $\mathbf{s}\text{, plusLog, minusLog}\leftarrow$\textit{quicksort4trim($\mathbf{s}$)} \newline\Comment{\hspace{3.5cm} sort by first sub-elements}
        \For{$\forall q \in$ minusLog }
            \State $\mathcal{A} -= \mathbf{D}_q^T\mathbf{D}_q$
        \EndFor
        \For{$\forall q \in$ plusLog }
            \State $\mathcal{A} += \mathbf{D}_q^T\mathbf{D}_q$
        \EndFor
        \State $\boldsymbol{\theta} \leftarrow \mathcal{NS}(\mathcal{A})$
        \State update scores in $\mathbf{s}$
    \EndWhile
    
    \State align control points and return $\mathbf{R}$ and $\mathbf{t}$
\EndProcedure

\end{algorithmic}
\end{algorithm}

\subsection{Incremental UPnP}

\textit{REPPnP} solves for a linear nullspace and therefore relies on an algebraic cost function. Geometrically optimal solvers can return superior results, but require the closed-form solution of a non-linear objective. Such an alternative for the camera resectioning problem is given by the \textit{UPnP} algorithm~\cite{kneip14}. In simple terms, \textit{UPnP} expresses the sum of geometric object space errors (i.e. point-to-ray distances) as a polynomial function of the quaternion parameters of the exterior orientation of the camera. This energy is then minimized in closed-form by finding the roots of the first-order optimality conditions using a Gr\"obner basis solver. Interestingly, the algorithm also employs accumulations over the input correspondences to generate the values of the elimination template, which is why our fast trimming strategy may also be applied to this non-linear objective. However, the equations of the original paper need to be slightly reformulated in order to single out a clear function of accumulators that can be updated incrementally. We do this here for the central case, but the rule can easily be extended to the non-central case as well. For details, the reader is kindly referred to~\cite{kneip14}.

Let $\boldsymbol{\theta}$ be a four-vector of the quaternion variables. The rotation of the 3D point into the camera frame is given by $\mathbf{R}(\boldsymbol{\theta})\mathbf{p}_q$, which---for the sake of a simplified derivation---is rewritten as the product
\begin{equation}
    \mathbf{R}(\boldsymbol{\theta})\mathbf{p}_q = \boldsymbol{\Phi}(\mathbf{p}_q)_{3\times 10} \cdot \mathbf{m}(\boldsymbol{\theta})_{10\times 1}.
\end{equation}
$\boldsymbol{\Phi}(\mathbf{p}_i)$ is a $3\times 10$ matrix that is filled with elements of $\mathbf{p}_i$, and $\mathbf{m}(\boldsymbol{\theta})$is a $10\times 1$ vector filled with all order-2 forms of the quaternion variables. The object-space error is given by
\small
\begin{equation}
    \boldsymbol{\eta}_q = (\mathbf{f}_q\mathbf{f}_q^T - \mathbf{I}) \left[ \boldsymbol{\Phi}(\mathbf{p}_q) + \mathbf{H}^{-1} \sum_{i=1}^{N} w_i \left[ \mathbf{f}_i\mathbf{f}_i^T - \mathbf{I} \right] \Phi(\mathbf{p}_i) \right] \mathbf{m}(\boldsymbol{\theta})
\end{equation}
\normalsize
where
\begin{equation}
    \mathbf{H}=\sum_{i=1}^{N}w_i (\mathbf{I}-\mathbf{f}_{i}\mathbf{f}_{i}^{T}),
\end{equation}
and $\mathbf{I}$ is the $3\times$ identity matrix. $w_i=1$ if correspondence $i$ is considered. The overall objective energy as a function of individual accumulators is finally given by
\small
\begin{equation}
    E = \sum_{i=1}^{N} \boldsymbol{\eta}_i^T \boldsymbol{\eta}_i =  \mathbf{m}(\boldsymbol{\theta})^T \mathcal{A}  \mathbf{m}(\boldsymbol{\theta})\text{, where}
\end{equation}
\footnotesize
\begin{equation}
    \mathcal{A} = \mathbf{A}_1 + \mathbf{A}_2\mathbf{H}^{-1}\mathbf{A}_0 + \mathbf{A}^T\mathbf{H}^{-1}\mathbf{A}_2^T + \mathbf{A}^T\mathbf{H}^{-1}\mathbf{A}_3\mathbf{H}^{-1}\mathbf{A}\nonumber
\end{equation}
\normalsize
\begin{eqnarray}
  \mathbf{A}_0 & = & \sum_{i=1}^{N}w_i (\mathbf{f}_i\mathbf{f}_i^T-\mathbf{I})\Phi(\mathbf{p}_i) \nonumber \\
  \mathbf{A}_1 & = & \sum_{i=1}^{N}w_i \left[\Phi(\mathbf{p}_i)\right]^T(\mathbf{f}_i\mathbf{f}_i^T-\mathbf{I})^2\Phi(\mathbf{p}_i) \nonumber \\
  \mathbf{A}_2 & = & \sum_{i=1}^{N}w_i \left[\Phi(\mathbf{p}_i)\right]^T(\mathbf{f}_i\mathbf{f}_i^T-\mathbf{I})^2 \nonumber \\
  \mathbf{A}_3 & = & \sum_{i=1}^{N}w_i (\mathbf{f}_i\mathbf{f}_i^T-\mathbf{I})^2. \nonumber
\end{eqnarray}

Again, we solve this problem using our incremental sorting and accumulation algorithm $\textit{quicksort4trim}$. Rather than summing up all terms for which $w_i \neq 0$, we register swaps across the $x$-th percentile boundary. The return variables \textit{plusLog} and \textit{minusLog} register terms for which $w_i$ toggles from 0 to 1 or 1 to 0, respectively, and only those terms need to be taken into account in order to update the accumulators $\mathbf{H}$, $\mathbf{A}_0$, $\mathbf{A}_1$, $\mathbf{A}_2$, and $\mathbf{A}_3$. Sorting is based on the reprojection error. The algorithm is summarized in Algorithm \ref{alg:incrupnp}.

\begin{algorithm}[t]
\caption{Robust Incremental UPnP}
\label{alg:incrupnp}
\begin{algorithmic}[1]

\Procedure{RobustUPnPIncr}{$[\{\mathbf{f}_1,\mathbf{p}_1\},\dots]$} 
    \State $\mathbf{H}=\sum_{i =1}^{N}(\mathbf{I}-\mathbf{f}_{i}\mathbf{f}_{i}^{T})$
    \State $\mathbf{A}_0 = \sum_{i=1}^{N} (\mathbf{f}_i\mathbf{f}_i^T-\mathbf{I})\Phi(\mathbf{p}_i)$
    \State $\mathbf{A}_1 = \sum_{i=1}^{N}\left[\Phi(\mathbf{p}_i)\right]^T(\mathbf{f}_i\mathbf{f}_i^T-\mathbf{I})^2\Phi(\mathbf{p}_i)$
    \State $\mathbf{A}_2 = \sum_{i=1}^{N} \left[\Phi(\mathbf{p}_i)\right]^T(\mathbf{f}_i\mathbf{f}_i^T-\mathbf{I})^2$
    \State $\mathbf{A}_3 = \sum_{i=1}^{N} (\mathbf{f}_i\mathbf{f}_i^T-\mathbf{I})^2$
    \State $\boldsymbol{\theta}_0 = \mathcal{GB}(\mathcal{A})$ \Comment{$\mathcal{GB}(\cdot)=$ UPnP solver}
    \State Solve for $\mathbf{t}$
    \State init $\mathbf{s}=[\{r_1,1\},\ldots,\{r_N,N\}]$ \Comment{$r_i$ denotes reprojection error for $i$-th feature}

    \State $\mathbf{H}=\sum_{i =1}^{\lfloor N/2\rfloor}(\mathbf{I}-\mathbf{f}_{i}\mathbf{f}_{i}^{T})$
    \State $\mathbf{A}_0 = \sum_{i=1}^{\lfloor N/2\rfloor} (\mathbf{f}_i\mathbf{f}_i^T-\mathbf{I})\Phi(\mathbf{p}_i)$
    \State $\mathbf{A}_1 = \sum_{i=1}^{\lfloor N/2\rfloor}\left[\Phi(\mathbf{p}_i)\right]^T(\mathbf{f}_i\mathbf{f}_i^T-\mathbf{I})^2\Phi(\mathbf{p}_i)$
    \State $\mathbf{A}_2 = \sum_{i=1}^{\lfloor N/2\rfloor} \left[\Phi(\mathbf{p}_i)\right]^T(\mathbf{f}_i\mathbf{f}_i^T-\mathbf{I})^2$
    \State $\mathbf{A}_3 = \sum_{i=1}^{\lfloor N/2\rfloor} (\mathbf{f}_i\mathbf{f}_i^T-\mathbf{I})^2$

    \While{not converged}
        \State $\mathbf{s}\text{, plusLog, minusLog}\leftarrow$\textit{quicksort4trim($\mathbf{s}$)} \newline\Comment{\hspace{3.5cm} sort by first sub-elements}
        \For{$\forall q \in$ minusLog }
            \State $\mathbf{H} -= \mathbf{I}-\mathbf{f}_{q}\mathbf{f}_{q}^{T}$
            \State $\mathbf{A}_0 -= (\mathbf{f}_q\mathbf{f}_q^T-\mathbf{I})\Phi(\mathbf{p}_q)$
            \State $\mathbf{A}_1 -= \left[\Phi(\mathbf{p}_q)\right]^T(\mathbf{f}_q\mathbf{f}_q^T-\mathbf{I})^2\Phi(\mathbf{p}_q)$
            \State $\mathbf{A}_2 -= \left[\Phi(\mathbf{p}_q)\right]^T(\mathbf{f}_q\mathbf{f}_q^T-\mathbf{I})^2$
            \State $\mathbf{A}_3 -= (\mathbf{f}_q\mathbf{f}_q^T-\mathbf{I})^2$
        \EndFor
        \For{$\forall q \in$ plusLog }
            \State $\mathbf{H} += \mathbf{I}-\mathbf{f}_{q}\mathbf{f}_{q}^{T}$
            \State $\mathbf{A}_0 += (\mathbf{f}_q\mathbf{f}_q^T-\mathbf{I})\Phi(\mathbf{p}_q)$
            \State $\mathbf{A}_1 += \left[\Phi(\mathbf{p}_q)\right]^T(\mathbf{f}_q\mathbf{f}_q^T-\mathbf{I})^2\Phi(\mathbf{p}_q)$
            \State $\mathbf{A}_2 += \left[\Phi(\mathbf{p}_q)\right]^T(\mathbf{f}_q\mathbf{f}_q^T-\mathbf{I})^2$
            \State $\mathbf{A}_3 += (\mathbf{f}_q\mathbf{f}_q^T-\mathbf{I})^2$
        \EndFor

        \State $\boldsymbol{\theta} = \mathcal{GB}(\mathcal{A})$
        \State Solve for $\mathbf{t}$
        \State update reprojection errors in $\mathbf{s}$
    \EndWhile
    
    \State return $\mathbf{R}$ and $\mathbf{t}$
\EndProcedure

\end{algorithmic}
\end{algorithm}

\begin{figure*}[t!]
\setlength{\abovecaptionskip}{-0.2cm}
\setlength{\belowcaptionskip}{-0.5cm}
    \begin{subfigure}{0.25\textwidth}
      \includegraphics[width=1\linewidth]{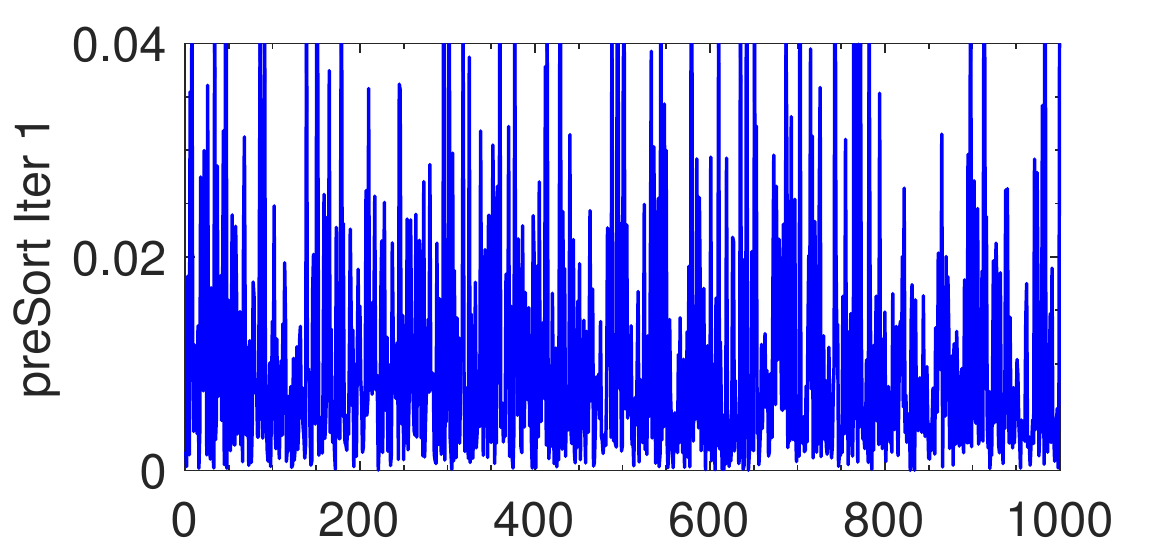}
    \end{subfigure}   %
    \hspace{-0.1 in}
    \begin{subfigure}{0.25\textwidth}
      \includegraphics[width=1\linewidth]{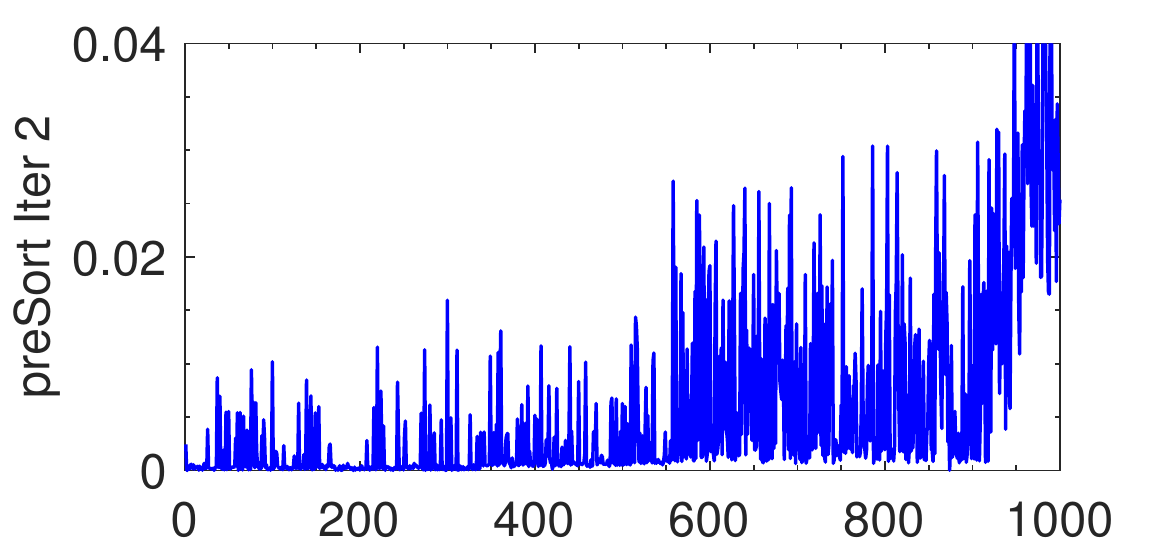}
    \end{subfigure}
        \hspace{-0.1 in}
        \begin{subfigure}{0.25\textwidth}
      \includegraphics[width=1\linewidth]{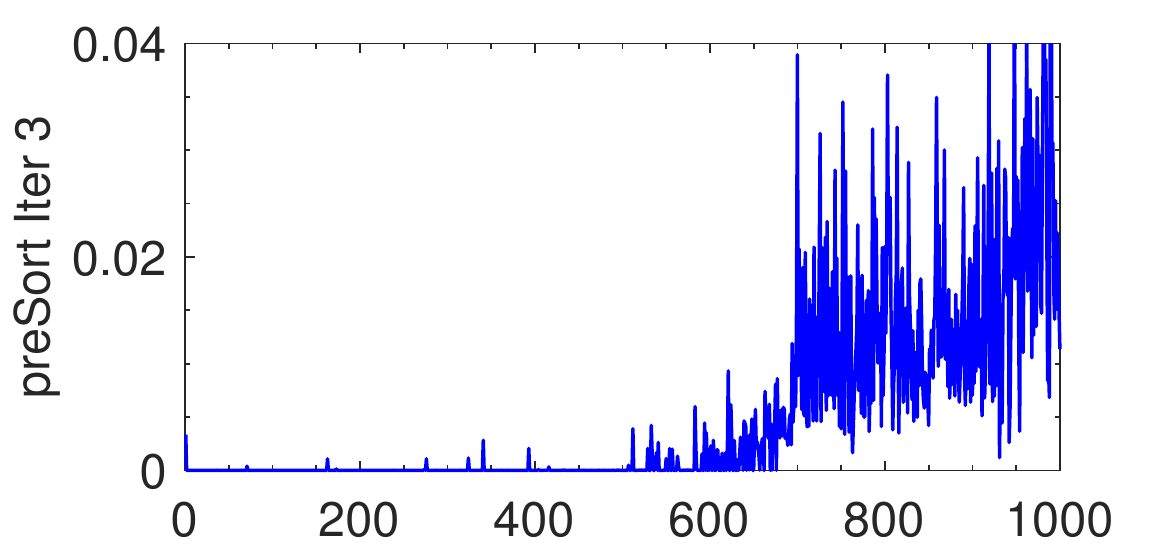}
    \end{subfigure}
        \hspace{-0.1 in}
            \begin{subfigure}{0.25\textwidth}
      \includegraphics[width=1\linewidth]{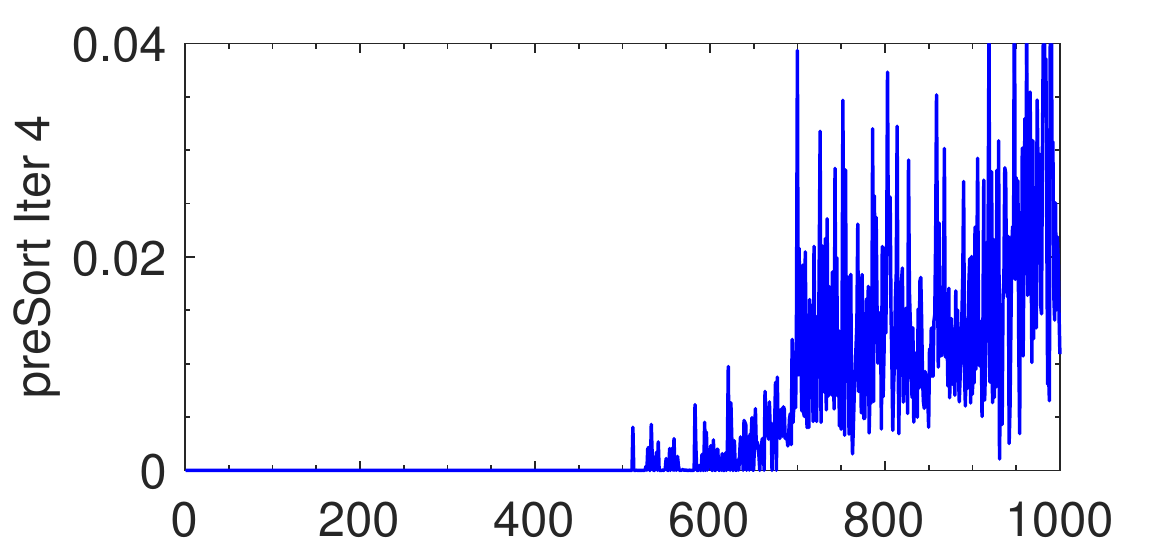}
    \end{subfigure}
    
        \begin{subfigure}{0.25\textwidth}
      \includegraphics[width=1\linewidth]{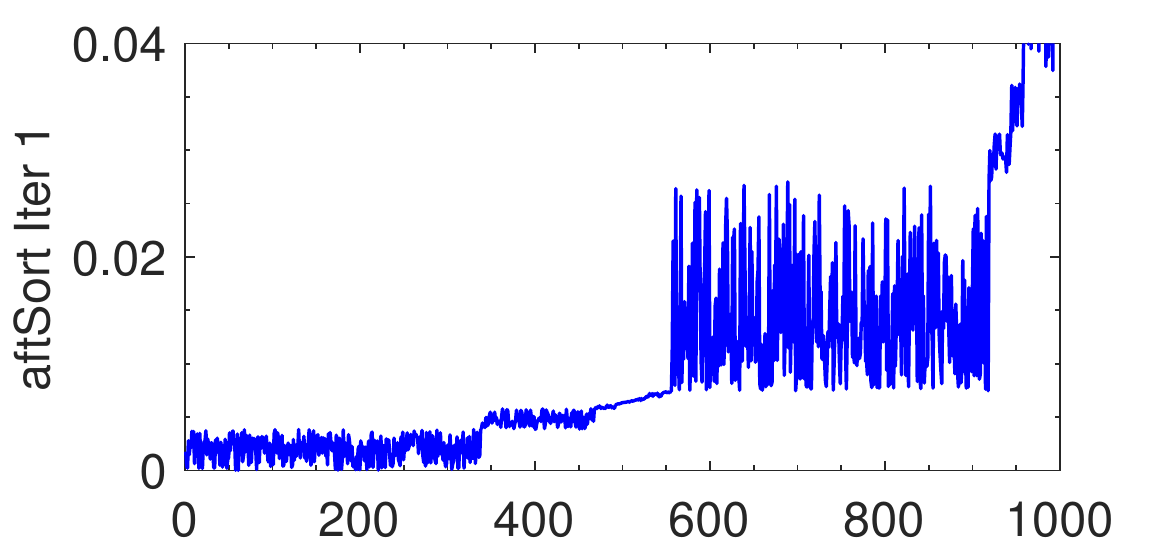}
        \label{fig: Score-Aft: Iter 1}
    \end{subfigure}   %
            \hspace{-0.1 in}
    \begin{subfigure}{0.25\textwidth}
      \includegraphics[width=1\linewidth]{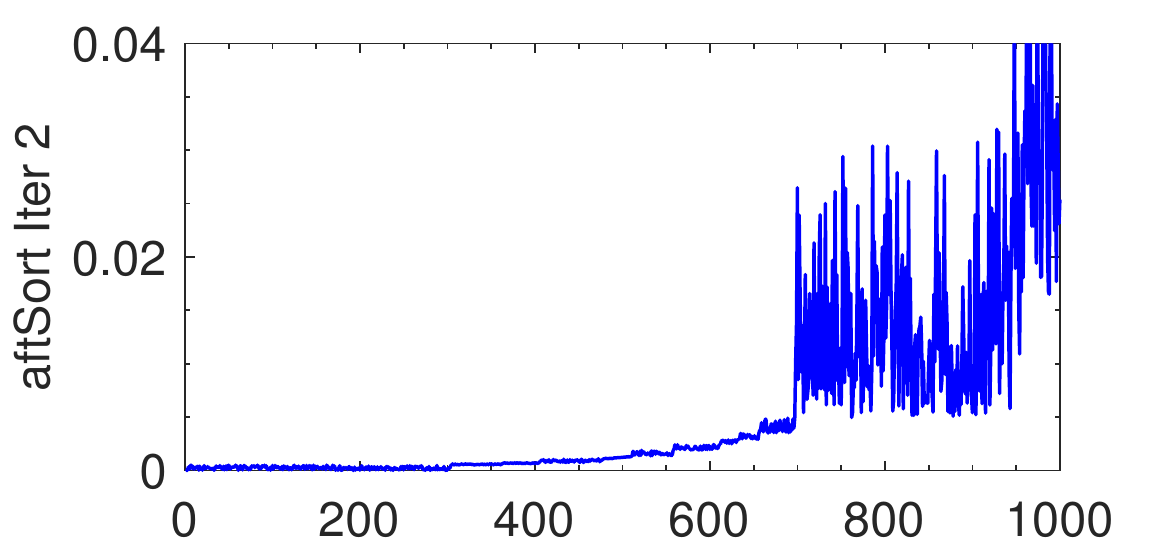}
        \label{fig: Score-Aft: Iter 2}
    \end{subfigure}
                \hspace{-0.1 in}
        \begin{subfigure}{0.25\textwidth}
      \includegraphics[width=1\linewidth]{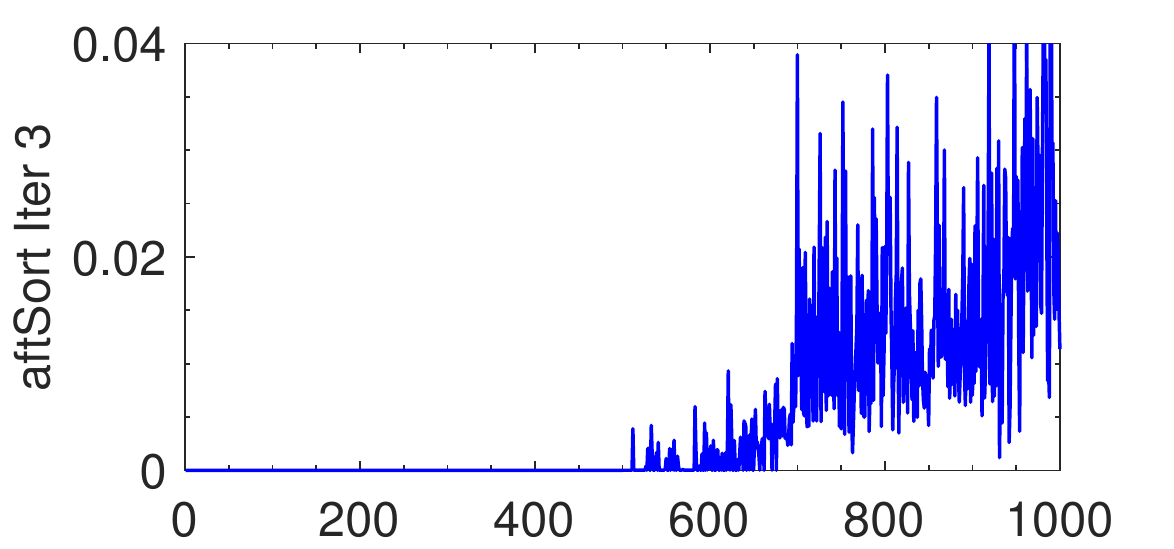}
        \label{fig: Score-Aft: Iter 3}
    \end{subfigure}
                \hspace{-0.1 in}
            \begin{subfigure}{0.25\textwidth}
      \includegraphics[width=1\linewidth]{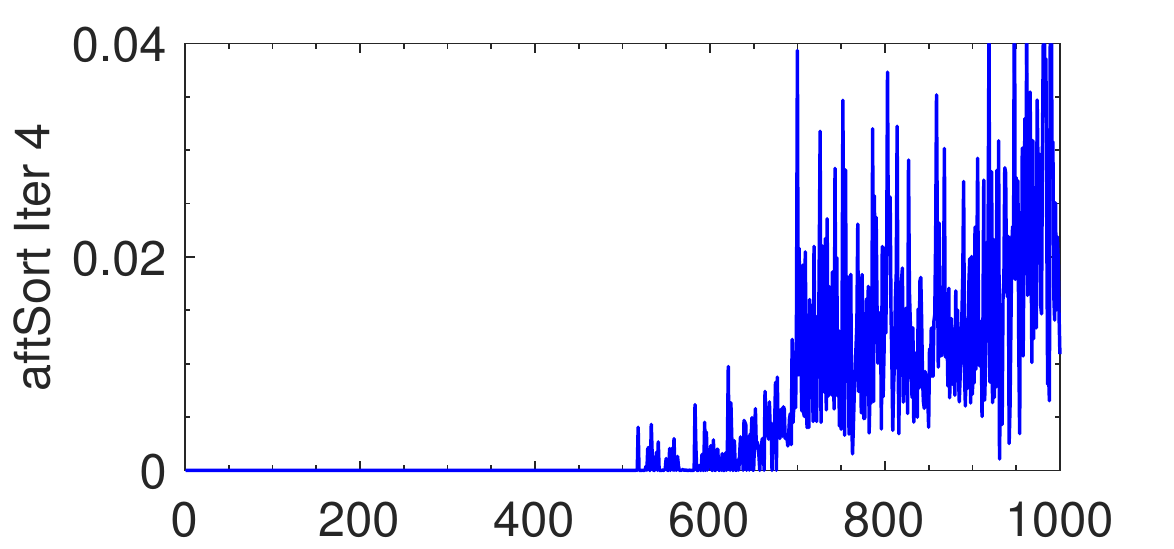}
        \label{fig: Score-Aft: Iter 4}
    \end{subfigure}
    \caption{
\label{fig:Scores}
Residual errors $\mathbf{s}$ in each iteration before (top) and after (bottom) sorting. The results are obtained for one example using our proposed \textit{RobustUPnPincr} algorithm, a robust variant of the \textit{UPnP}~\cite{kneip14} algorithm employing partial incremental trim fitting.}
\end{figure*}

\section{Experimental Results}

We compare our incremental implementations of \textit{REPPnP}~\cite{ferraz14} and \textit{UPnP}~\cite{kneip14} against their corresponding algorithms employing full sorting and naive accumulation as well as two state-of-the-art, non-robust PnP methods, i.e., \textit{ePnP}~\cite{lepetit09} and Ransac based on the P3P algorithm~\cite{kneip11} (denoted \textit{P3PRansac}). All results are obtained by C++ implementations running on an Intel® Core™ i5-8250U 8-core CPU clocked at 1.60GHz × 8.

We conduct rigorous synthetic experiments assuming a virtual calibrated camera with a focal length of 800. We generate random 2D-3D correspondences by distributing 3D points in a [-2 2] $\times$ [-2 2] $\times$ [4 8] volume in front of the camera. We finally add different levels of uniform noise to the image measurements and produce random outliers in the data by randomizing the direction vectors of a fraction of the correspondences. Ground-truth rotations and translations are generated and used to transform the 3D points into the world frame. Absolute errors in rotation (in rad) and translation (in m) are calculated and compared for different numbers of correspondences, noise levels and outlier fractions. If $\mathbf{R}_{gt}$ and $\mathbf{R}$ denote the ground truth and estimated rotations, our error is given by $\|\mathbf{R}^T\mathbf{R}_{\text{gt}}-\mathbf{I}\|_{\text{Fr}}$, which is equivalent to the angle of the residual rotation expressed in radiants. We furthermore evaluate computational efficiency by running each set of experiments more than 1000 times and considering the average running time.

\subsection{Computational Efficiency}

The goal of this paper is to present an algorithmic approach to improve the efficiency of robust trim-fitting in outlier-affected geometric regression problems. The computational efficiency comprises two parts. The first one is given by the efficiency of the sorting itself, which has been analysed in Figure \ref{fig:Figure1}. Here we focus on the second part, which is the impact of reducing the number of operations required during accumulation.

\begin{figure}[!b]
\setlength{\abovecaptionskip}{-0.0cm}
\setlength{\belowcaptionskip}{-0.3cm}
\centering
\includegraphics[scale = 0.42]{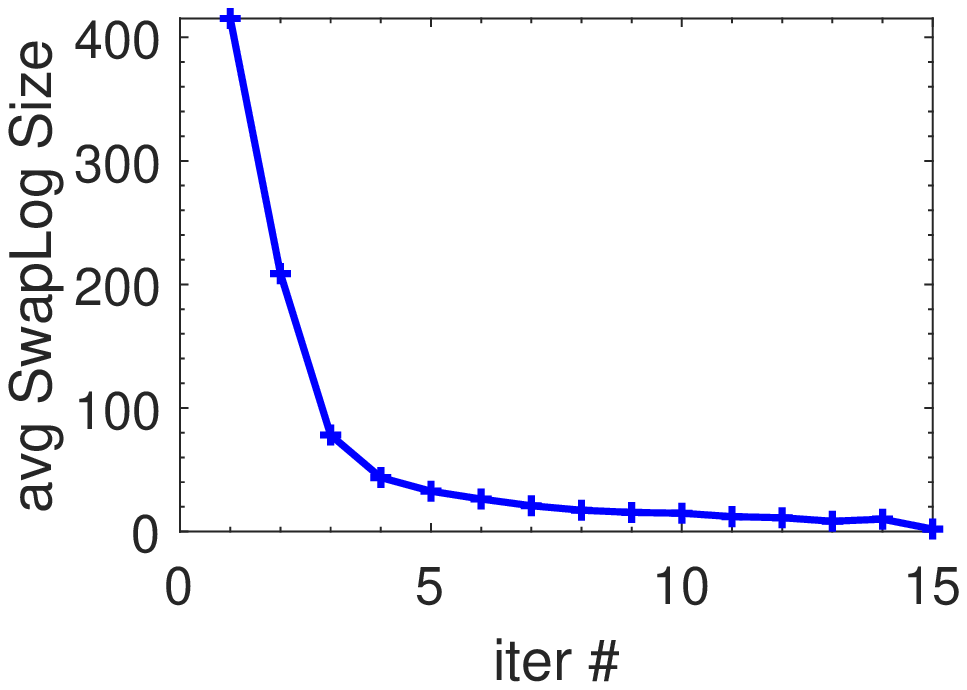}
\caption{Average number of non-redundant cross-percentile-boundary swaps at different iterations. The total number of correspondences is 2000.}
\label{fig: plot_avgSwapElement}
\centering
\end{figure}

Figure \ref{fig:Scores} illustrates partial incremental sorting in operation for an experiment of \textit{RobustUPnPIncr}. It visualizes the score values before (top row) and after (bottom row) sorting. Assuming that the amount of outliers is less than 50\%, a fixed 50 percent ($x$=50 in (\ref{eq:trimFitting})) threshold is used throughout all experiments in this work. As illustrated, the perturbed presort scores become partially ordered after sorting. Furthermore, the residual errors within the 50-th percentile are becoming gradually smaller, thus indicating convergence of the algorithm. It can furthermore be seen that the scores before and after sorting present decreasing differences as iterations proceed, thus indicating that sorting and accumulation efficiency increase during convergence. The average number of non-redundant cross-percentile-boundary swaps in each iteration is illustrated in Figure \ref{fig: plot_avgSwapElement}, confirming a fast decline over the very first iterations.

\begin{figure}[b!]
\setlength{\abovecaptionskip}{-0.2cm}
\setlength{\belowcaptionskip}{-0.3cm}
\begin{subfigure}{0.24\textwidth}
\flushleft
\includegraphics[width=1\textwidth]{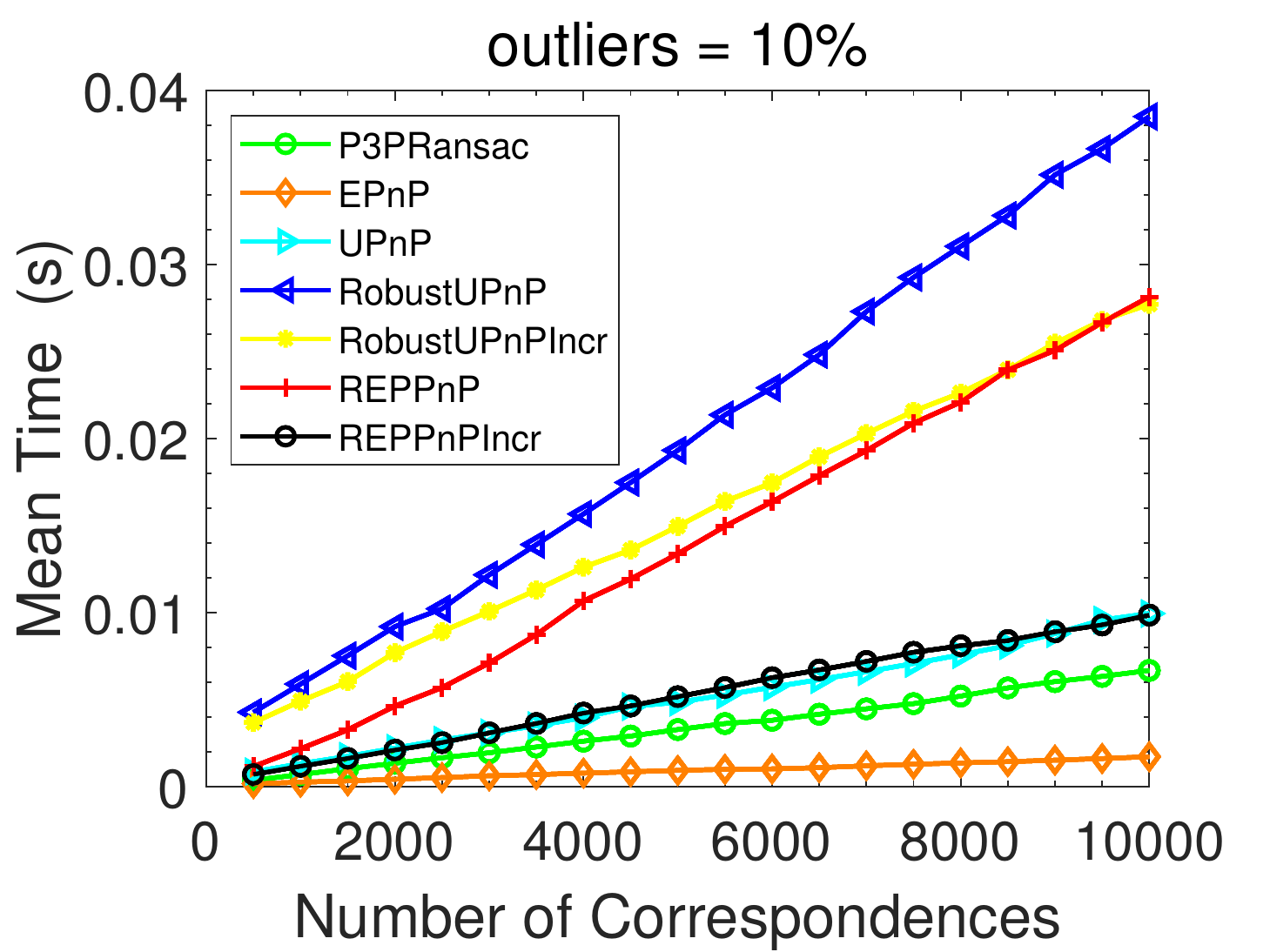}
\label{fig: outlier 10 percent}
\end{subfigure}
        \hspace{-0.1 in}
\begin{subfigure}{0.24\textwidth}
\includegraphics[width=1\textwidth]{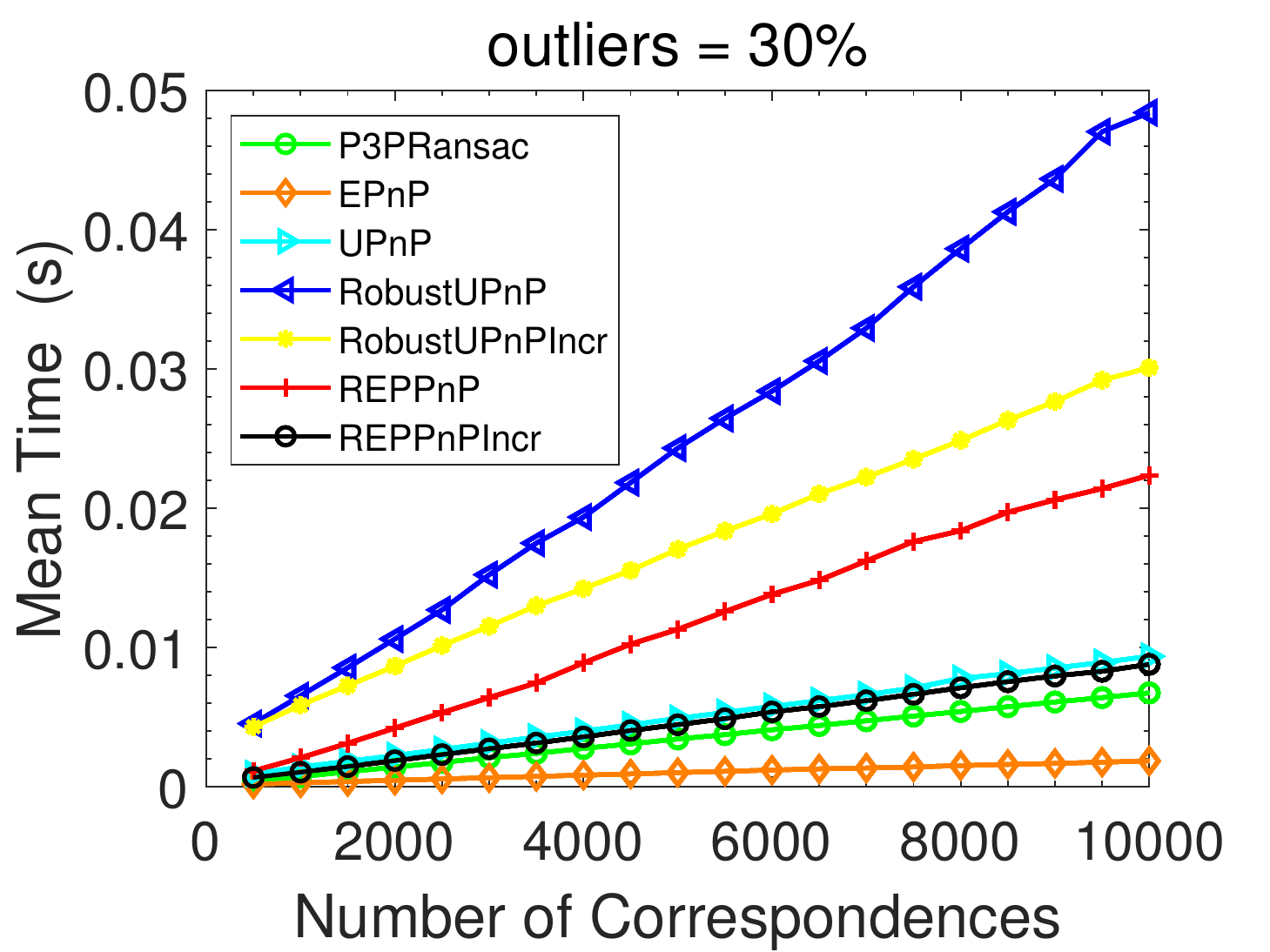}
\label{fig: outlier 30 percent}
\end{subfigure}
\caption{Average execution time for different number of correspondences and outliers.}
\label{fig:runningTime}
\end{figure}

Given that the linear complexity step of the employed solvers outweighs all other steps (at least for sufficiently many points), this behavior leads to a substantial increase in computational efficiency. We evaluate the mean running time of \textit{P3PRansac}~\cite{kneip11}, \textit{ePnP}~\cite{lepetit09}, \textit{REPPnP}~\cite{ferraz14}, \textit{REPPnPIncr} (\textit{REPPnP} + fast geometric trim fitting, Algorithm \ref{alg:incrreppnp}), \textit{UPnP}~\cite{kneip14}, \textit{RobustUPnP} (\textit{UPnP} + regular geometric trim fitting) and \textit{RobustUPnPIncr} (\textit{UPnP} + fast geometric trim fitting, Algorithm \ref{alg:incrupnp}). The results are summarized in Figure \ref{fig:runningTime}, where the computational efficiency is evaluated for a varying number of correspondences. It is highly interesting to see that \textit{REPPnPincr} becomes at least twice as fast and achieves a running time that is comparable to \textit{P3PRansac}. All experiments are executed for constant Gaussian noise of 3 pixels and with a fixed outlier fractions of 10\% (left) and 30\% (right).

\subsection{Number of Correspondences and Noise}

\setlength{\abovecaptionskip}{-0.2cm}
\setlength{\belowcaptionskip}{-0.5cm}
\begin{figure}[t!]
\begin{subfigure}{0.24\textwidth}
\flushleft
\includegraphics[width=1\textwidth]{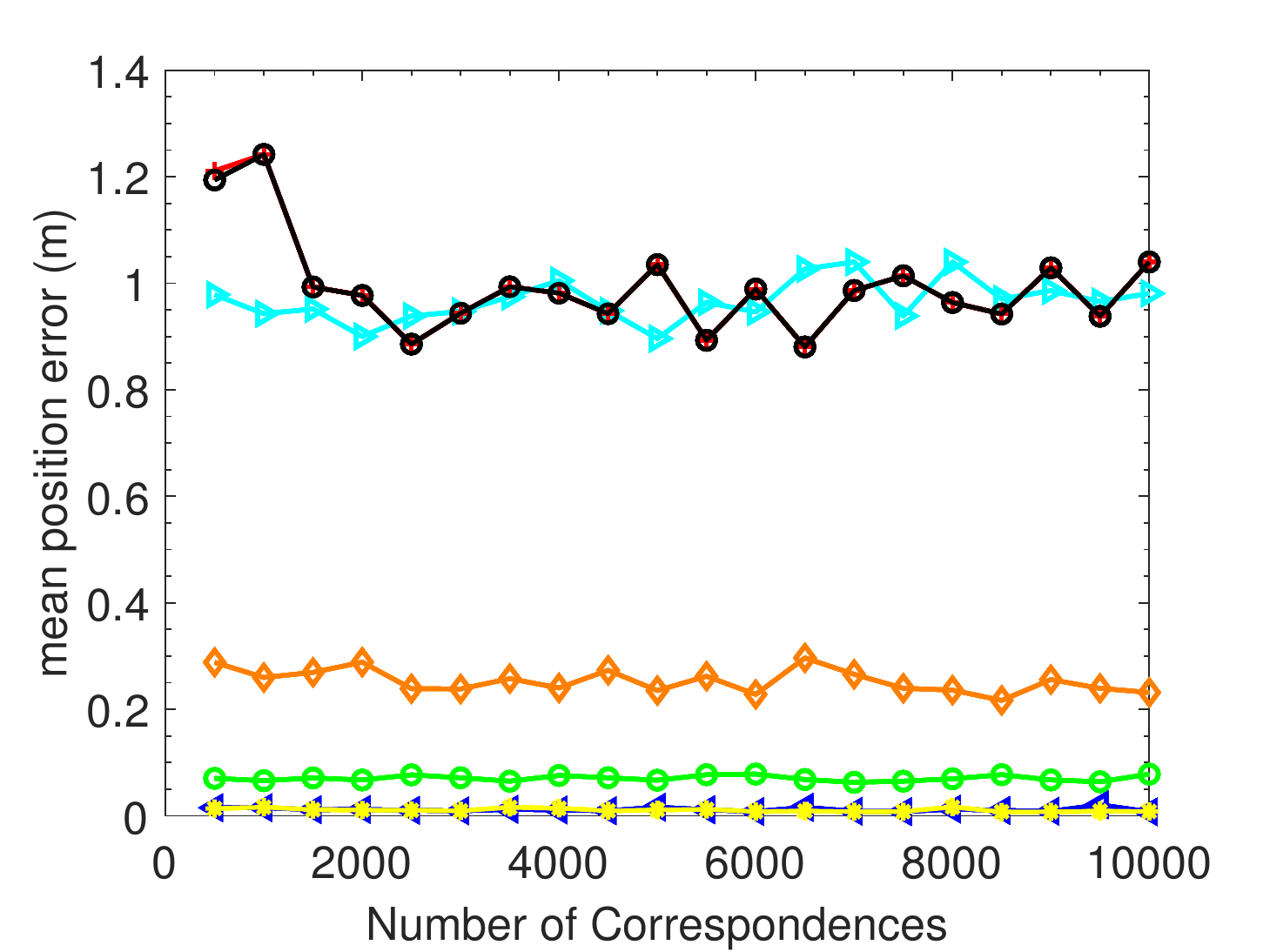}
\end{subfigure}
        \hspace{-0.1 in}
\begin{subfigure}{0.24\textwidth}
\includegraphics[width=1\textwidth]{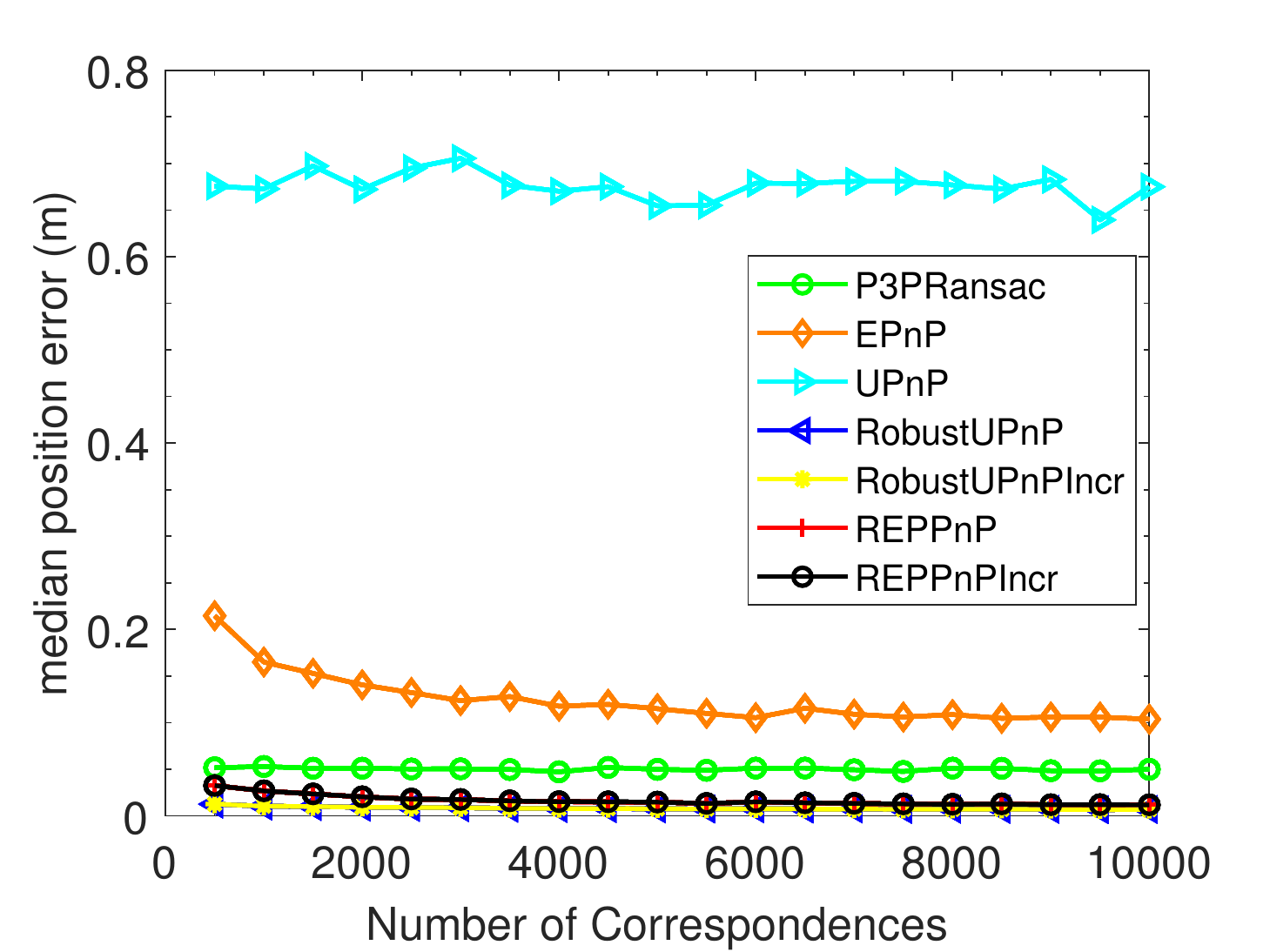}
\end{subfigure}

\begin{subfigure}{0.24\textwidth}
\flushleft
\includegraphics[width=1\textwidth]{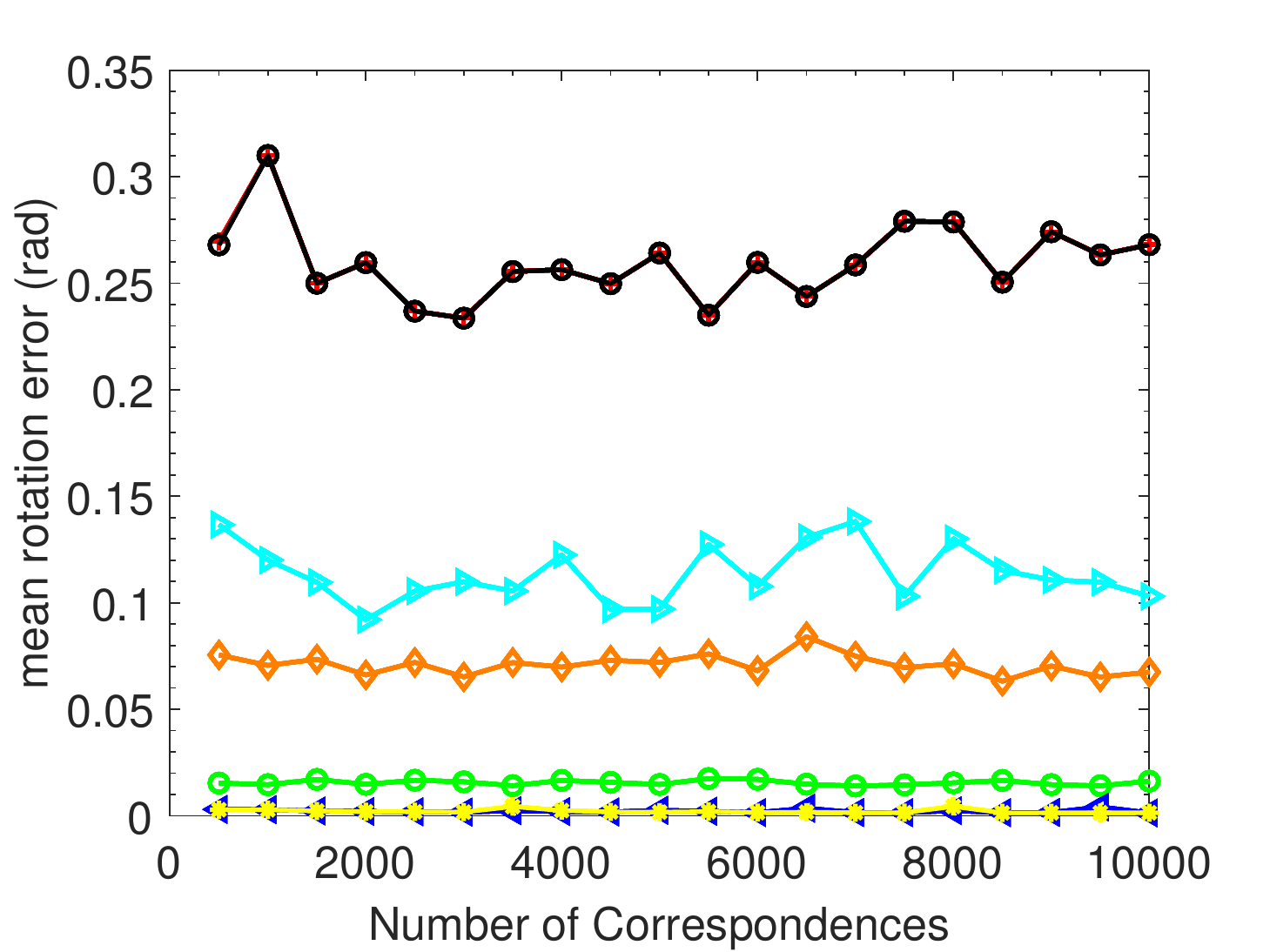}
\label{fig: numCorres mean rotation error}
\end{subfigure}
        \hspace{-0.1 in}
\begin{subfigure}{0.24\textwidth}
\includegraphics[width=1\textwidth]{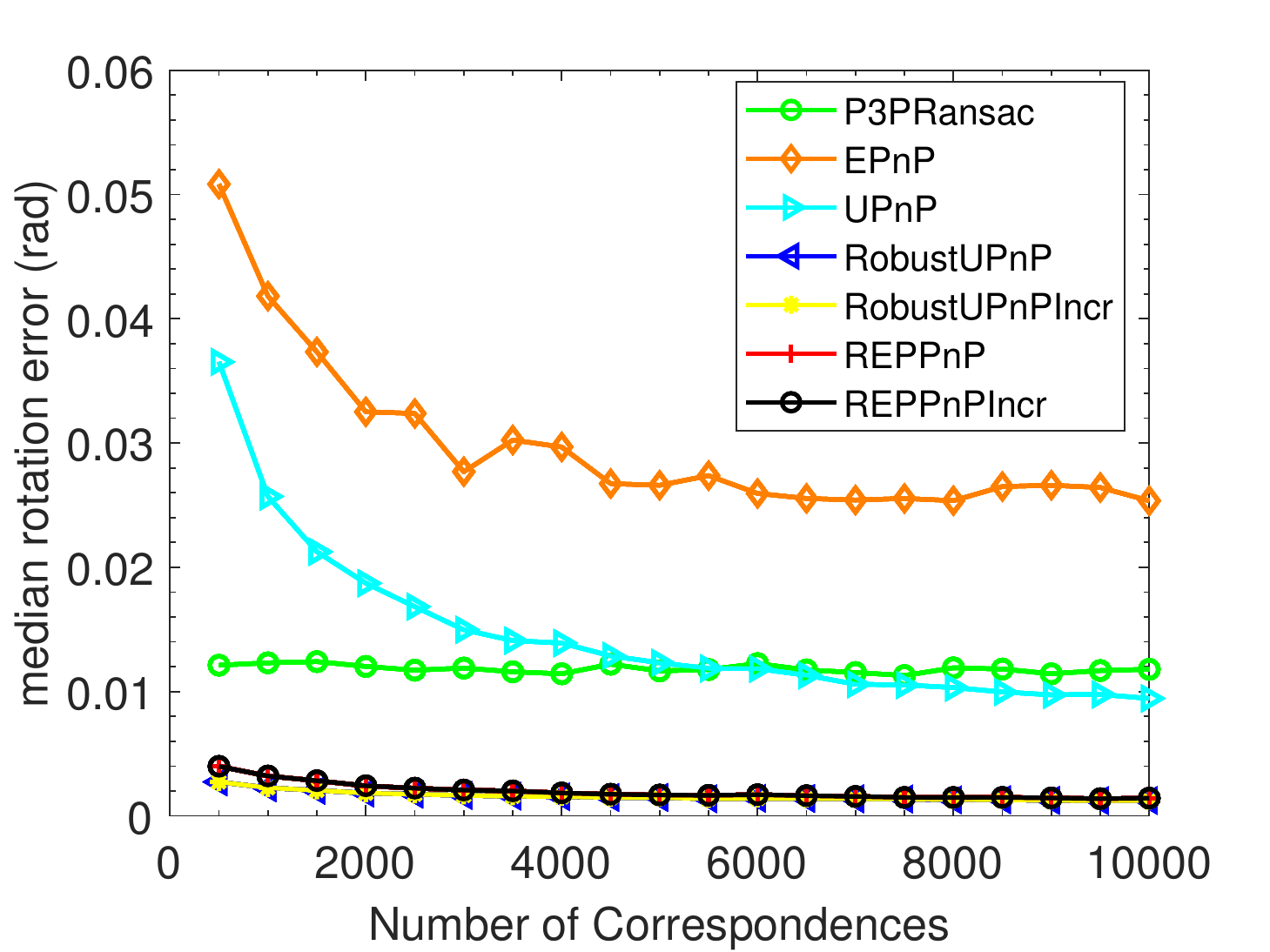}
\label{fig: numCorres median rotation error}
\end{subfigure}
\caption{Errors for all algorithms for a varying number of correspondences.}
\label{fig: Errors. Varying the number of correspondences.}
\end{figure}

Figure \ref{fig: Errors. Varying the number of correspondences.} shows the errors over the number of correspondences. Results are obtained for uniform noise of 3.0 pixels and an outlier fraction of 30\%. Both \textit{RobustUPnP} and \textit{RobustUPnPIncr} return much lower mean and median (position and rotation) errors than \textit{P3PRansac}, \textit{UPnP} and \textit{EPnP}, demonstrating simultaneous strong rejection of outliers and high accuracy of the trim-fitting based, geometric solver. Furthermore, note that---while the median errors of the algebraic and the geometric solvers are practically identical---\textit{REPPnP} and \textit{REPPnPIncr} are significantly outperformed by their geometrically optimal counter-parts in terms of the mean error owing to the fact that the algebraic solver often fails to converge. We have tried both the original implementation of~\cite{ferraz14} as well as our own re-implementation of the algorithm. The indicated results are the best results we were able to obtain using the algebraic error criterion. Note that the errors of \textit{RobustUPnP} and \textit{RobustUPnPIncr} are practically the same, which indicates that the incremental sorting merely increases computational efficiency without affecting the results. The same is true for \textit{REPPnP} and \textit{REPPnPIncr}.



\begin{figure}[t!]
\setlength{\abovecaptionskip}{-0.2cm}
\setlength{\belowcaptionskip}{-0.2cm}

\begin{subfigure}{0.24\textwidth}
\flushleft
\includegraphics[width=1\textwidth]{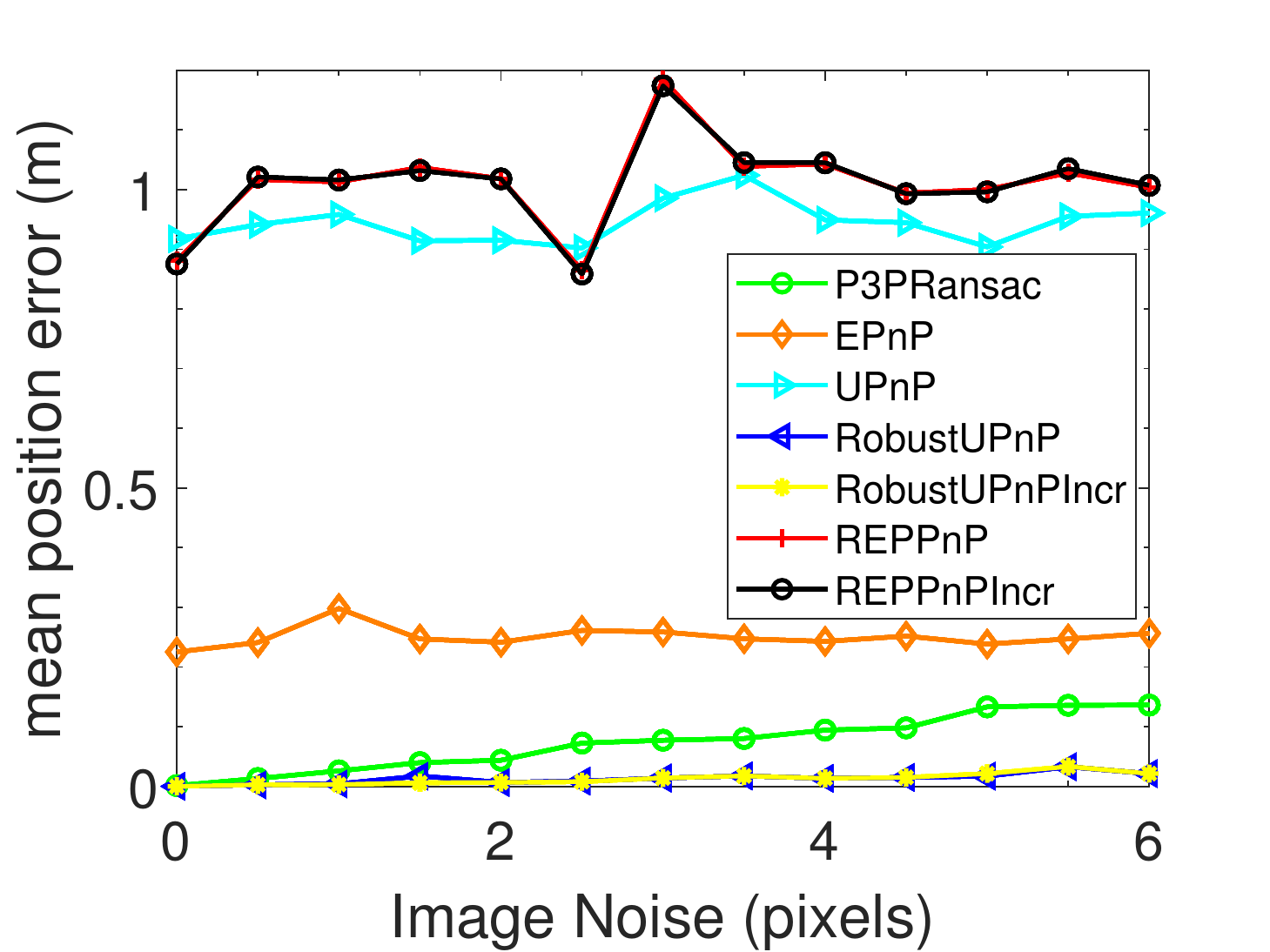}
\end{subfigure}
\hspace{-0.1 in}
\begin{subfigure}{0.24\textwidth}
\includegraphics[width=1\textwidth]{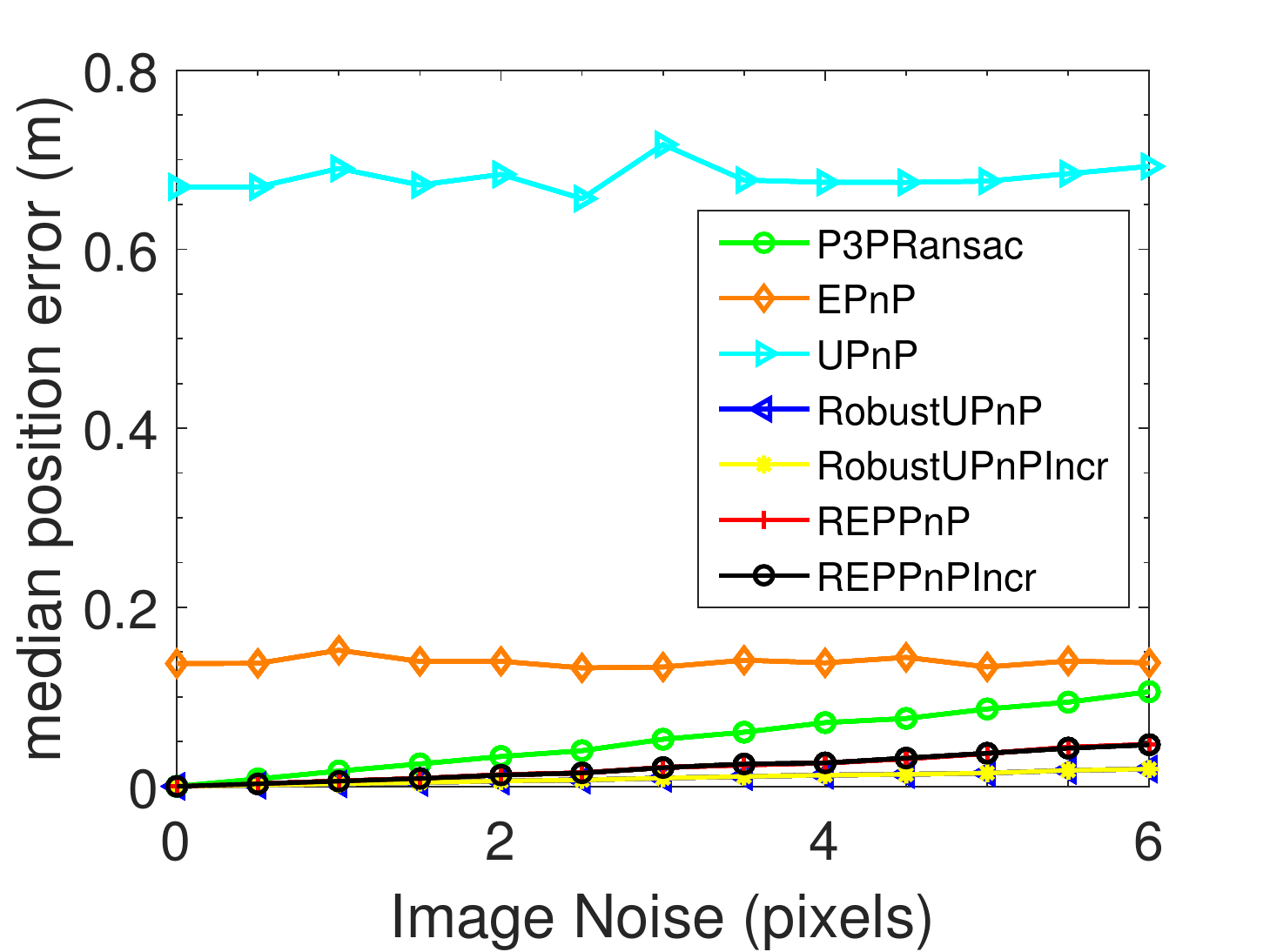}
\end{subfigure}
\begin{subfigure}{0.24\textwidth}
\flushleft
\includegraphics[width=1\textwidth]{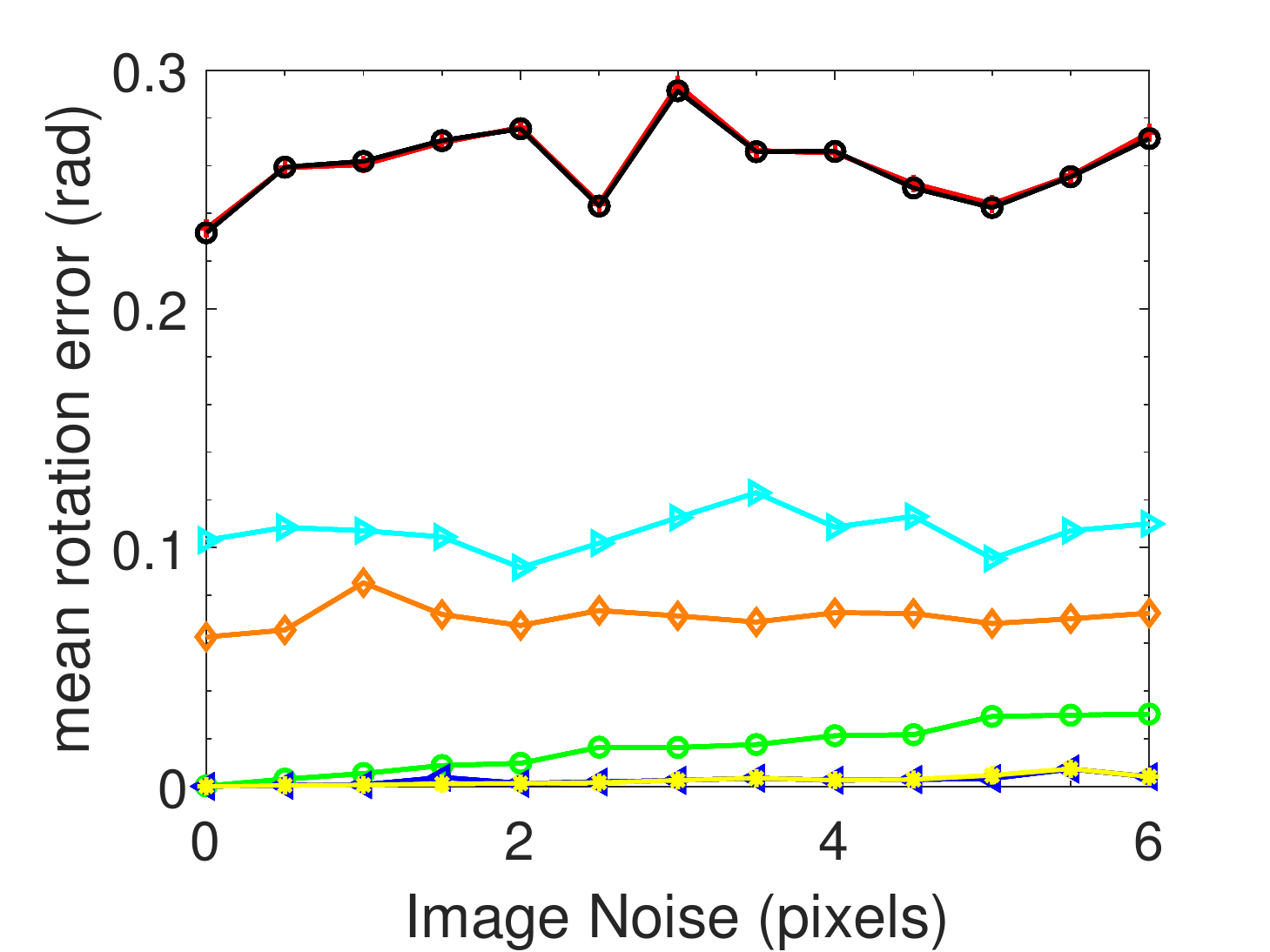}
\label{fig: noiseExperiment mean rotation error}
\end{subfigure}
\hspace{-0.1 in}
\begin{subfigure}{0.24\textwidth}
\includegraphics[width=1\textwidth]{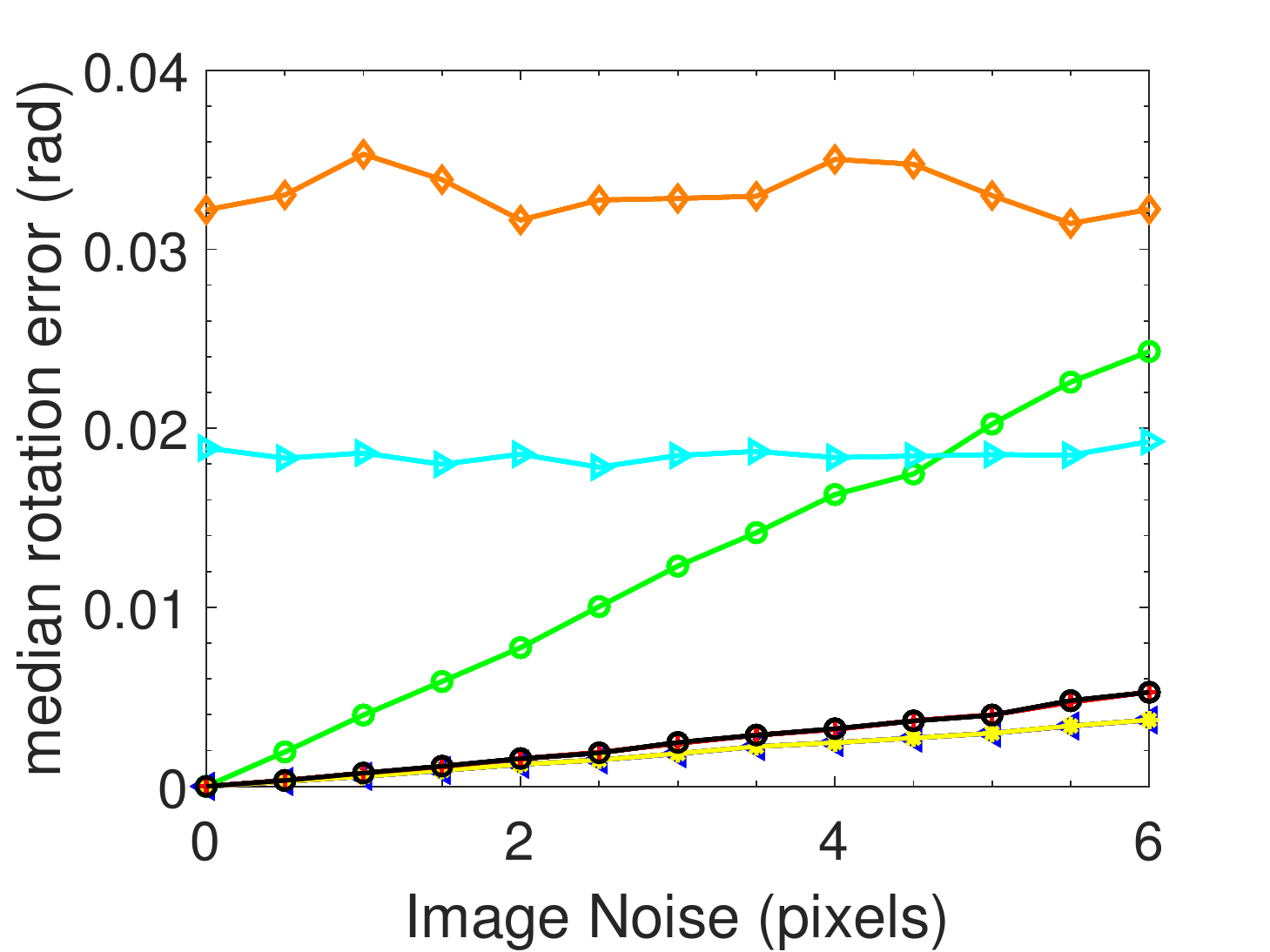}
\label{fig: noiseExperiment median rotation error}
\end{subfigure}

\caption{Errors for varying noise levels.}
\label{fig: Errors. Varying the noise level.}
\end{figure}

Figure \ref{fig: Errors. Varying the noise level.} finally shows errors for varying uniform noise levels reaching from 0 to 6.0 pixels. The experiments use an outlier fraction of 30\% and 2000 correspondences. The result demonstrates that \textit{REPPnP} and \textit{RobustUPnP} (with and without fast incremental trim fitting) have lower median position and rotation errors than \textit{P3PRansac}, \textit{EPnP} and \textit{UPnP}, and the geometric solver ultimately produces the most accurate results.

\subsection{Robustness against outliers}

\begin{figure}[b!]
\setlength{\abovecaptionskip}{-0.2cm}
\setlength{\belowcaptionskip}{-0.15cm}

\begin{subfigure}{0.24\textwidth}
\flushleft
\includegraphics[width=1\textwidth]{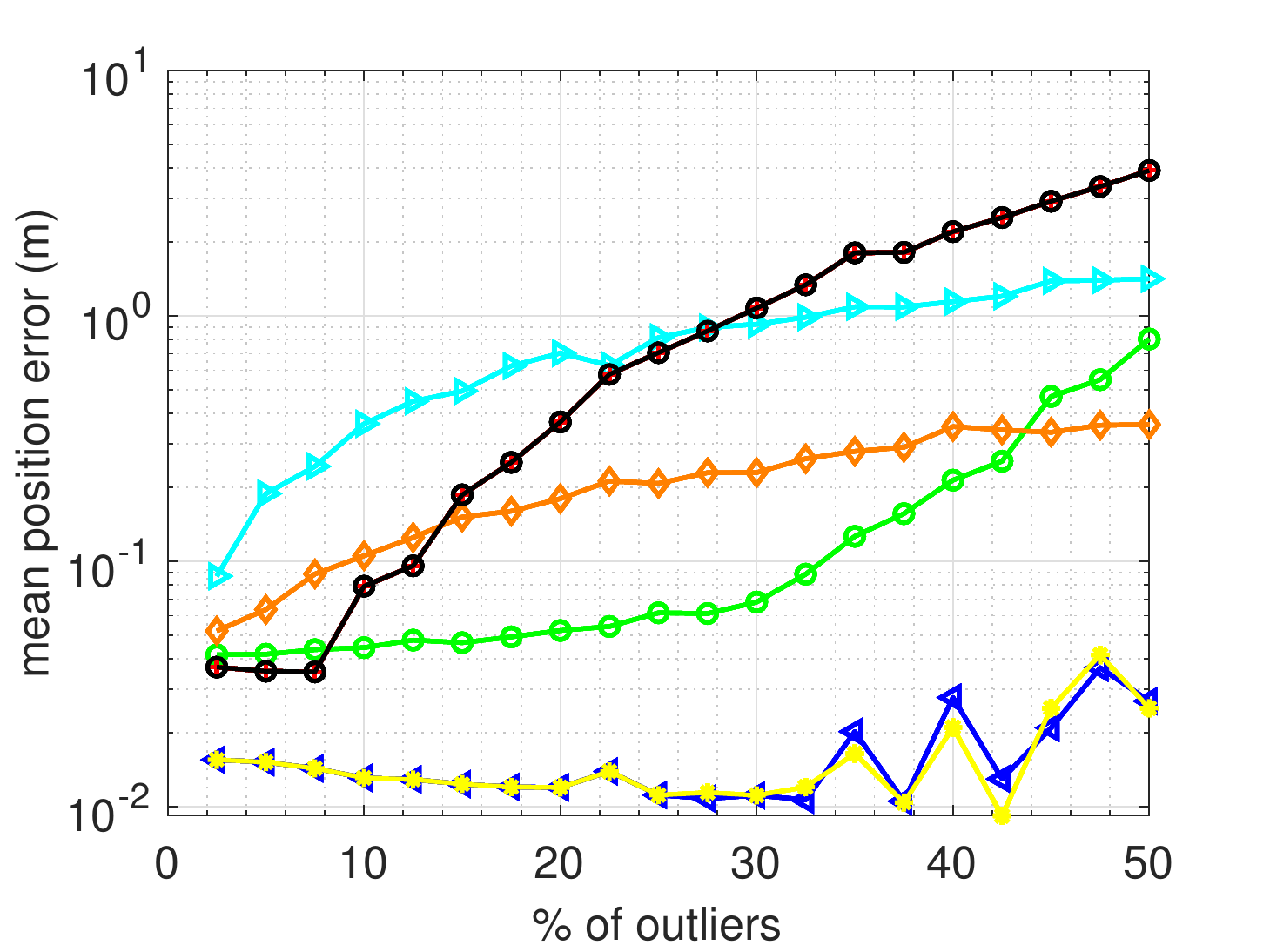}
\label{fig: outlierExperiment mean position error}
\end{subfigure}
%
%
\hspace{-0.1 in}
\begin{subfigure}{0.24\textwidth}
\flushleft
\includegraphics[width=1\textwidth]{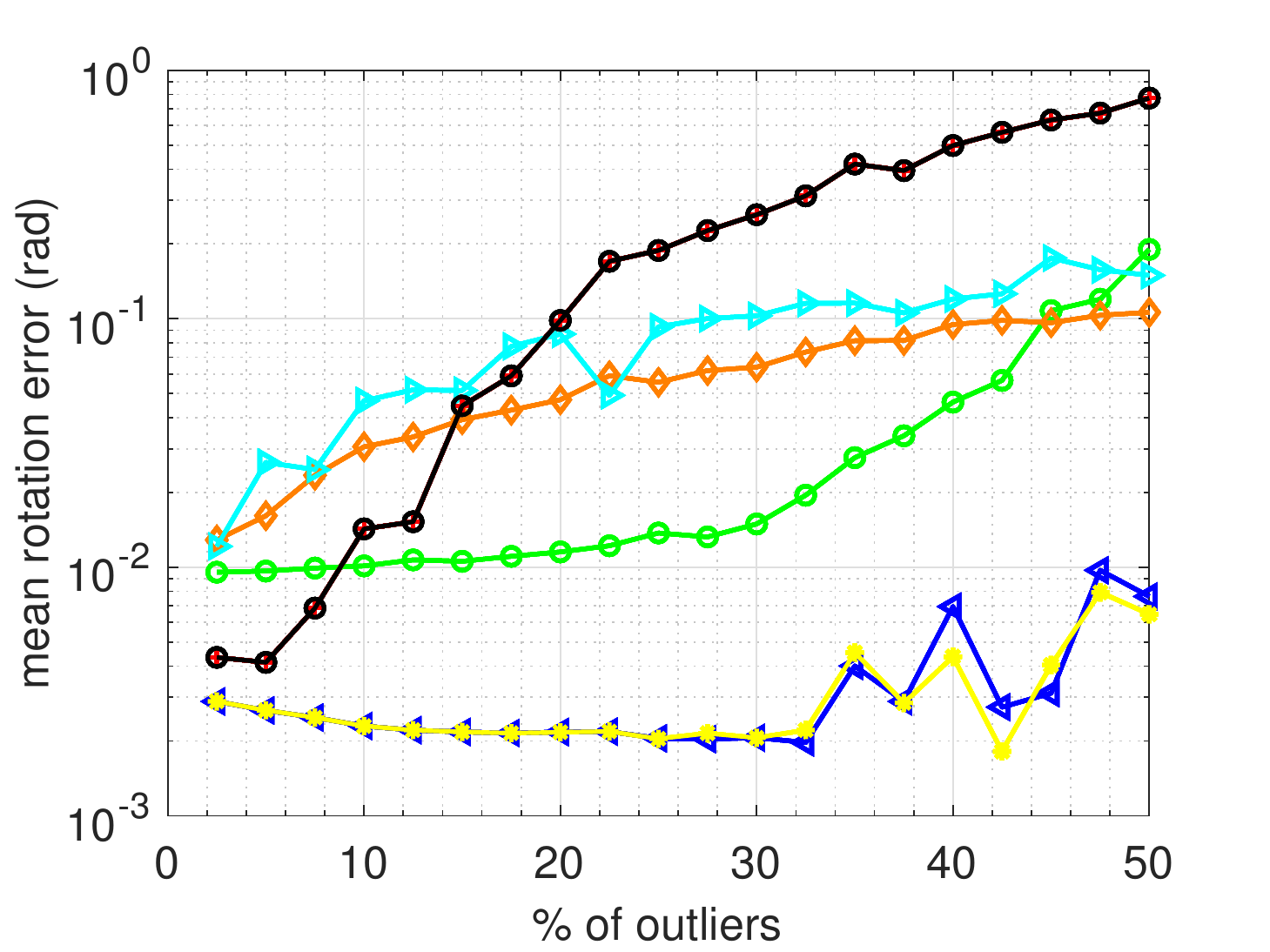}
\label{fig: outlierExperiment mean rotation error}
\end{subfigure}
%

\caption{Errors for varying outlier fractions. Note that labels are identical with previous figures.}
\label{fig: Errors. Varying the value of outlier fraction.}
\end{figure}

Figure \ref{fig: Errors. Varying the value of outlier fraction.} finally shows errors obtained for varying outlier fractions between 2.5\% and 50\%. The noise level is kept at 3.0 pixels and the number of correspondences remains 2000.
%
%
As can be observed in Figure \ref{fig: Errors. Varying the value of outlier fraction.}, an increasing outlier fraction leads to increasing mean errors for \textit{EPnP} and \textit{UPnP}, which is natural owing to their non-robust nature. \textit{P3PRansac} has a break down point of about 30\%, while the algebraic solver shows high instability starting from outlier levels as low as 10\%. \textit{RobustUPnP} and \textit{RobustUPnPincr} demonstrate the best performance and have a similar break-down point than \textit{P3PRansac}, but lower errors owing to the geometric nature of the algorithm.


\section{Conclusion}

The presented algorithm makes an astute use of the internal swapping operations in trimming methods for a significant reduction of computation time of the most important, linear-complexity step in geometry solvers. We have furthermore demonstrated that this technique is amenable to non-linear geometrically optimal solvers. This leads to a significant improvement in success rate compared to linear algebraic null space solvers, and makes trim-fitting a viable alternative in practical applications. The present work limits the evaluation to camera resectioning, for which very good performance is achieved, but the application of Ransac followed by a refinement over the inlier subset remains the gold standard. Our current investigations focus on higher dimensional problems, for which the proposed technique could achieve an unprecedented mix of accuracy, success rate and computational efficiency.

%
%

{\small

\bibliographystyle{IEEEtran}
}

\end{document}